\RequirePackage{fix-cm}
\documentclass[smallextended]{svjour3}       
\smartqed  
\usepackage{amsmath}
\usepackage{amsfonts}
\usepackage{graphicx}
\usepackage{graphics} 
\usepackage[colorlinks,linkcolor=blue,citecolor=magenta]{hyperref}
\usepackage{subfigure} 
\usepackage{color}
\usepackage{marvosym}
\usepackage{wasysym}
\usepackage{tikz}
\begin{document}

\title{A Projective Simulation Scheme for Partially-Observable Multi-Agent Systems 
}
\subtitle{}


\author{R. Kheiri
}


\institute{R. Kheiri\at
              Isfahan University of Technology, Department of Physics, Isfahan, 84156-83111, Iran\\
              \email{r.kheiry@ph.iut.ac.ir}           
}

\date{Received: date / Accepted: date}

\maketitle
\definecolor{mygreen}{rgb}{0.0, 0.5, 0.0}

\begin{abstract}
We introduce a kind of partial observability to the projective simulation (PS) learning method. It is done by adding a belief projection operator and an observability parameter to the original framework of the efficiency of the PS model. I provide theoretical formulations, network representations, and situated scenarios derived from the invasion toy problem as a starting point for some multi-agent PS models.   
\end{abstract}
\section{Introduction}
Even though the experimental quantum computing has been difficult to achieve, theoretical quantum artificial intelligence and machine learning, as a branch, has been flourishing in the recent years \cite{Dunjko2,Biamonteie}. In short, and as a wish, quantum computing might affordably speed up our classical coding. Furthermore, since every quantum computation can have a classical counterpart, we may build the classical models and think of their quantum apparitions. In this sense, projective simulation (PS), as a new model for classical and quantum artificial intelligence, needs to be expanded in all AI directions as a lot of works have been done so far. Particularly, some classical studies are \cite{Briegel1,Briegel2,Briegel3,Alexey1} and some quantum works could be \cite{Briegel1,Dunjko1,Tiersch11,JensClausen.,Alexey2}.

In a classical point of view, the former \cite{Briegel1} is the original introduction of the PS learning method through the compositional memories as well as the initial episodic memory of a given agent. Other studies, \cite{Briegel2,Briegel3,Alexey1}, can be considered as some expansions of the model in the existing fields of AI. Among them, temporal correlations, associative memory and comparison with some other learning methods as Q-learning and Learning Classifier System (XCS) have been discussed in ref. \cite{Briegel2}, and meta-learning or learning to learn has been introduced in ref. \cite{Briegel3}. In the latter in turn, \cite{Alexey1}, the generalization of some similar but distinct stimuli has been studied. As other important branches, partially observability and multi-agent approaches have been considered widely in artificial intelligence \cite{KaelblingLittman,book:Russel,book:Sigaud} and reinforcement learning \cite{Fujita,Doshi} which we are about to introduce a first step in the PS model via the current study.    

A single agent in a partially observable environment may be unable to observe the current state completely where the notion of "belief state" represents the agent's current belief about the current "world state". In this approach, a fully observable environment is just a special case of the partial observability when the belief state ($b$) is equal to the current percept ($s$) for every so-called world percept $s$ at all times; therefore, partially observable artificial intelligence (AI), as a generalization, can add more realistic examples to the entirely observable AI scenarios \cite{Actingoptimally,book:Russel}.  

In a multi-agent partially observable approach, despite the controversial issue of what is a multi-agent setting\footnote{\begin{quote}
\emph{``Indeed, even the
seemingly simpler question--What is a (single) agent?--has resisted a definitive
answer. For our purposes, the following loose definition will suffice: Multiagent systems are those systems that include multiple autonomous entities with either diverging information or diverging interests, or both.''} \cite{book:MultiagentShoham}
\end{quote}}, in addition to the current world state, the parameters of the other agents also may not be always observable for which there are some mainstreams in the literature, such as interactive partially observable \cite{Gmytrasiewicz1,Gmytrasiewicz2,Panella} and decentralized partially observable \cite{VictorLesser,Bernstein1,bookchapter:Oliehoek,Amato1} subfields.
Decentralized headlines consider local observations in a cooperative game with a common (joint) reward function. In this treatment, the belief states are about the configuration of other agents as well as the world states where the optimal solutions are computed centrally for all of the agents. This category is then related, correspondingly, to if the system state is jointly fully observable (by all of the agents involved) or not \cite{book:Sigaud}. Likewise, in the interactive partially observable frameworks, the belief states include beliefs about the other agents as well as the physical environment \cite{Gmytrasiewicz2}; however, in contrast to a decentralized solution, here the formulation is applied for self-interest agents and individual solutions. In short, if a situation were locally observable by the agents regarding a common reward function for a cooperative task, the classification would be a decentralized approach. Alternatively, in case agents make a decision for individual self-interest games, the method is going toward building an interactive model, giving rise to equilibrium points and other game-theoretical issues. In both conditions, a generalized belief state is often needed for the optimal solutions. 

 
Another option which can be considered to a multi-agent learning viewpoint is communication between the involving agents. Obviously, learning and communication can be related to each other from learning to communicate to communication as learning \cite{bookWeiss}. Exchanging information, on the one hand, might increase the common-interests in a cooperative task. On the other hand, communication can be just for self-interests since, for instance, whenever an agent starts to help, another agent may compensate it and look forward to the next round from which both make use of a long-term communication to have a better individual efficiency for the separated tasks. There could be lots of other criteria and assumptions as well.\footnote{for example see \cite{MatthijsSpaan} as a concise review of multi-agent models in the partially observable environments including the communication classifications withal.}

The context of multi-agent systems is quite interdisciplinary. It has been used in economics to system biology and evolution, and from social science to software engineering and robotics. Nevertheless, in the quantum domain, we can refer to some related studies as distributed quantum computation \cite{Buhrmanh,Perseguersk} and quantum game theory \cite{Miakkisz,Psiotrowlski,fFlitney}. In this paper, after mentioning some basic principles of projective simulation, we try to familiarize the PS learning method with the notion of a partially observable environment. It is done by introducing a set of environmental belief states and also an observability parameter in the formulation of efficiency. As our situated examples, we use a multi-agent invasion setting where another agent as an interpreter teaches a portion of the world percepts to a defender where the system states remain jointly fully observable. Afterwards, every agent can be considered as a simultaneous learner-teacher wherein there could be some self-interested aspects too. To have interactive models as well as the decentralized frameworks, however, we look forward to having further studies including a more general belief state. We express our evaluations classically here, though our formulation might be modified to work also in the quantum domain. 
\section{Projective Simulation}
\label{intro}
Projective simulation, to put it briefly, is an embodied (situated) learning method which representing a reductionist approach of a brain-like learning (thinking) scheme by utilizing random walks between some network of clips via their interface edges as episodic and compositional memories. In the article titled ``projective simulation for artificial intelligence'' \cite{Briegel1}, the authors formulate the PS model using a toy problem as "invasion game" and checked the speed of learning, maximum blocking efficiency, etc. Here, we expand our situated examples on it for the early step of introducing partial observability and multi-agent contexts to the PS network. 

As a fundamental desire, the PS learning process is considered to be decoupled from immediate motor action since the random walks happen between the virtual (fictitious) percept \textcircled{\textit{s}} and action \textcircled{\textit{a}} clips in the memory (networks of clips) itself which can be modified (updated); this occurs both in the number of clips and in the transition probabilities between the clips via reflections or some compositional properties before the real action $a$ takes place. Due to the assumptive capability of the content modification or creation of new clips in the network of clips,\footnote{I may also think of annihilation of old useless clips or the decay of some portion of the network (clips or edges) as a result of some neurological disorder.} one may think of every percept clip \textcircled{\textit{s}} or every action clip \textcircled{\textit{a}} just as a network clip \textcircled{\textit{c}} in the system. For this reason, one can formulate the PS model just on the network' clips \textcircled{\textit{c}} irrespective of the kind of them. Therefore, the transitional probabilities between every two clips $c_i$ and $c_j$ in the time step $t$ can be written as a normalized conditional probability function built of the wight transitions ${\omega}^{(t)} (c_i ,c_j )$.  
\begin{equation}
P^{(t)} ( c_j |c_i ) = \frac{{\omega}^{(t)} (c_i, c_j )}{\sum_k {\omega}^{(t)} (c_i, c_k)}.
\label{eqprob}
\end{equation}
Where ${\omega}^{(t)} (c_i ,c_j ) = f (h^{(t)} (c_i ,c_j ) )$ is modified when the edges $(c_{k_l}, c_{m_l})$ are traversed during the last random walk as an adaptation rule of
\begin{equation}
h^{(t+1)} (c_i, c_j ) = h^{(t)} (c_i, c_j ) - \gamma ( h^{(t)} (c_i, c_j ) -1 ) + \sum_l \delta (c_i , c_{k_l} ) \delta (c_j, c_{m_l}) {\lambda}^{(t+1)}
\label{eqtrans}
\end{equation}
, where $0\leq \gamma \leq 1$ is a forgetting factor (damping parameter), and $\lambda$ is a non-negative reward incrementing the related \textit{h}-value function. It follows that the forgetting factor $ \gamma $ in Eq.~\ref{eqtrans} can show a positive effect on speed-up learning in a changing environment and an adverse effect on the amount of efficiency.\footnote{The probabilities in projective simulation change fractionally as do in the fictitious play models, and the role of the forgetting (dissipation) factor in the projective simulation can be indirectly compared with the task of the discounted reward or learning factor in Q-learning \cite{WatkinsDayan}, or the step size parameter in the linear update scheme of reinforcement learning \cite{Verbeeck}.} The simplest function for $\omega$ is for $f(h)=h$, as used in the current study as well as previous studies.\footnote{Though there could be other alternative functions like an exponential function, as mentioned in \cite{Alexey1}, similar to that of original reinforcement learning.} In addition, the weight matrix $\omega = h^{(t)} (c_i, c_j)$ is initially unit ($h^{(1)} (c_i, c_j )=1$) for all edges.

Subsequently, learning happens by changing the transition probabilities of Eq.~\ref{eqprob} on a given pair of $(c_i, c_j )$ by updating the rule of Eq.~\ref{eqtrans}. 
One can evaluate the efficiency of learning, $r^{(t)}$, by adding the amounts of the desirable joint probabilities $P^{(n)} (a_s^{*} , s)$ at time $t$, where $ a_s^{*} $ is a pleasing action on a given percept $ s $.
Assuming that
\begin{equation}
r^{(t)} = \sum_s P^{(t)} (s)\, P^{(t)} (a_s^{*} |s) , \quad P^{(n)} (s) = \frac{1}{\textit{number of different percepts}},
\end{equation}   
then we have
\begin{equation}
r^{(t)} = \frac{1}{N} \sum_{i=1}^N P^{t} (a_{s_i}^{*} | s_i ).
\label{eqoriggin}
\end{equation} 

There are other important properties in the PS scheme such as the edge-glow mechanism $ g(c_i, c_j ) \in [0, 1] $ \cite{Briegel2,Alexey1}, which refers to a bypass reward of $ g^{(t+1)} (c_i, c_j ) \, \lambda^{(t+1)} $ for a non-rewarded edge $ (c_i, c_j ) $, where $ \lambda $ refers to a certain reward for the next-edge $ (c_j, c_k ) $. In other words, if an edge is rewarded, the previous edge can be strengthened by a temporal correlation. In this note, however, our examples contain only the invasion game within which one agent in a fully observable environment performs optimally in case we do not have such temporal correlations, where $ g=0 $ \cite{Briegel3}. Moreover, in the Appendix.~\ref{erageffredg}, we show that using the edge glow between two agents will destroy the partial observability in the multi-agent setting. Accordingly, the adaptation rule of Eq.~\ref{eqtrans} is enough for our purpose.

In the current study, we use Dirac notation, which is a useful theoretical method in both classical AI and quantum contexts, to show the influence of a partially observable environment on the efficiency of a given agent, though my results are presented classically here. 

\section{Method}
\label{soskssk} 
\subsection{A two-agent model}
\begin{figure*}
  \includegraphics[width=0.2\textwidth]{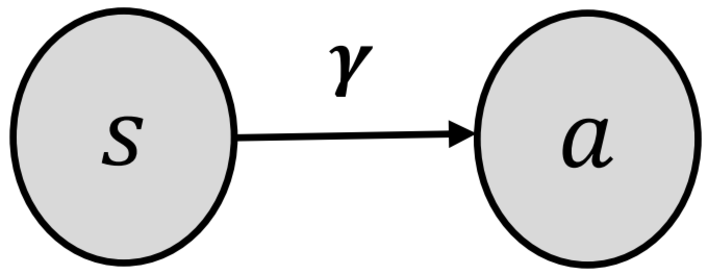}
\caption{The simplest clips network representation, according to Eq.\ref{eqfullyy}, where $ \gamma $ is the damping parameter in the adaptation rule (Eq.~\ref{eqtrans}) of projective simulation}
\label{plainpaldj}       
\end{figure*}

\begin{figure*}
  \includegraphics[width=0.33\textwidth]{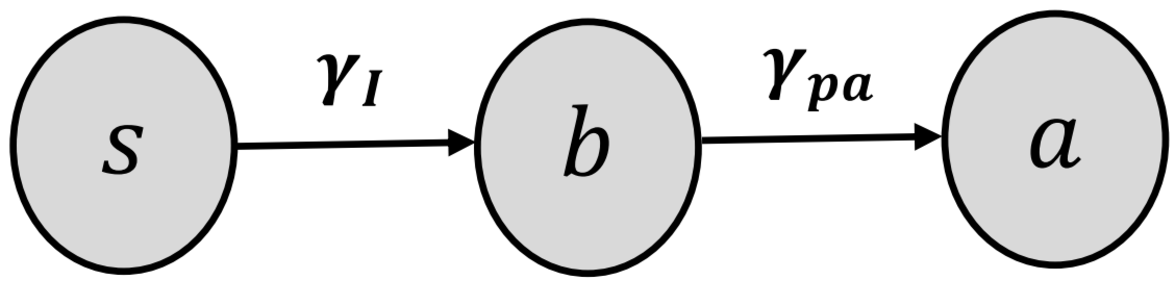}
\caption{The clips network representation according to Eq.\ref{blockeff}, where $ \gamma_{pa} $ and $ \gamma_{I} $ stand for the damping parameters of the protagonist agent and the interpreter, respectively. A protagonist agent here is an agent who does a domain action and an interpreter stands for an agent who performs a communication action}
\label{plakajstm}       
\end{figure*}

In the PS context, by remembering the efficiency of Eq.\ref{eqoriggin} and using a vector notation for $N$ different percept-actions, then
\begin{equation}
r^{(t)} = \frac{1}{N} \sum_{i=1}^N \, \langle a_{s_i}^{*} | s_i \rangle .
\label{eqfullyy}
\end{equation}
Where the superscript $ t $ was omitted for the classical probabilities ($ \langle a | s  \rangle = P^{t} ( a | s) $). A clips network representation for such a formula can be as that shown in Fig.\ref{plainpaldj}.
A given stochastic environment, so far, has been fully observable since the agent observes the world percepts $ s_i $. However, people may think of expanding a vector space on a different basis to make some belief states from the world states $ | s_i \rangle $ by other agents. As a tangible example, consider another intelligent agent as an interpreter (an \textbf{intelligent projector}), as illustrated in Fig.~\ref{interpreter:1}. We can span our real states (world states) $ | s_i \rangle  $ on the belief states  $ | b_i \rangle  $ which are related to fictitious action clips \textcircled{\textit{b}} in a PS learning network performed by an interpreter (I) with regard to $ s_i $. One can write a belief projection operator \cite{book:Sakurai} $ B_j $ as 
\begin{equation}
B_j =  |  b_j \rangle \langle b_j |.
\end{equation}
Then
\begin{equation}
| s_i \rangle = \sum_j \, |  b_j \rangle \langle b_j | s_i \rangle \quad , \quad  \sum_j \, \langle b_j | s_i \rangle = 1.
\end{equation}
\begin{equation}
r^{(t)}= \frac{1}{N} \sum_{i=1}^N \sum_{j=1}^{N'} \, \langle a_{s_i}^{*} | \, B_j \, | s_i \rangle = \frac{1}{N} \sum_{i=1}^N \sum_{j=1}^{N'} \, \langle a_{s_i}^{*}  | b_j \rangle \, \langle b_j | s_i \rangle
\label{blockeff}
\end{equation}    
Here, $ {N'} $ stands for the number of possible belief states which, in the immediate examples, are equal to the number of world states $ N $. A network representation containing the belief clip \textcircled{\textit{b}} is illustrated in Fig.~\ref{plakajstm}. One can also write the belief states' vectors with respect to the world states using an inverse matrix for probabilities $\langle b_i | s_i \rangle $. However, in the probability matrix of $\langle b_k | s_k  \rangle $, the summation of elements on a row is equal to one $ \sum_j \, \langle b_j | s_i \rangle = 1 $, but it is not the case for the summation on a column $ \sum_j \, \langle b_i | s_j \rangle $. We can also assume that
\begin{equation}
{\delta}_{ij} = \langle s_i | s_j \rangle \, = \langle a_i | a_j \rangle \ = \langle b_i | b_j \rangle  
\end{equation}
where needed.
As a more general case, one can consider a combinational operator $ (S+B)_j $ which regards an \textbf{observability parameter} $ \alpha $ for a given environment that is some portion $ \alpha $ of all percepts to be fully observable for a protagonist agent, with the rest of it being partially observable. In other words, a given world percept is visible with the probability of $ \alpha $ and is partially visible with the probability of $1 - \alpha $. 
 
\begin{equation}
\alpha \, S_j + ( 1 - \alpha ) \, B_j =  \alpha \, | s_j \rangle \langle s_j | \, + \, ( 1 - \alpha ) \,   |  b_j \rangle \langle b_j | , 
\end{equation}
\begin{equation}
r^{(t)} = \frac{1}{N} \sum_{i=1}^N \sum_{j=1}^{N'} \, \langle a_{s_i}^{*} | \, \alpha \, S_j + ( 1 - \alpha ) \, B_j \, | s_i \rangle ,
\end{equation}
\begin{equation}
r^{(t)}= \frac{\alpha }{N} \sum_{i=1}^N \, \langle a_{s_i}^{*} | {s_i} \rangle \, + \, \frac{1 - \alpha }{N} \sum_{i=1}^N \sum_{j=1}^{N'} \, \langle a_{s_i}^{*} | b_j \rangle \, \langle b_j | s_i \rangle .
\label{kabli}
\end{equation}
 
Further, we could include the imaginary space to define our generic belief state $ | (S+B)_{j} \rangle $ as   
\begin{equation}
| (S+B)_{j} \rangle = \sqrt{\alpha} \, | s_{j} \rangle  + i \, \sqrt{1 - \alpha } \, | b_j \rangle ,
\label{eqktswmtt}
\end{equation}
\begin{equation}
(S+B)_j = \, | (S+B)_{j} \rangle \,\, \langle (S+B)_{j} | .
\label{thefifi}
\end{equation}
Then, one can define the efficiency of Eq.\ref{kabli} as the real part of $ r^{(t)} $, that is
\begin{equation}
r^{(t)}= \frac{1}{N} \sum_{i=1}^N \sum_{j=1}^{N'} \, \langle a_{s_i}^{*} | (S+B)_j | s_i \rangle
\end{equation}
\begin{eqnarray}
efficiency \equiv Real \, [ r^{(t)} ] = \frac{\alpha }{N} \sum_{i=1}^N \, \langle a_{s_i}^{*} | {s_i} \rangle \, + \, \frac{1 - \alpha }{N} \sum_{i=1}^N \sum_{j=1}^{N'} \, \langle a_{s_i}^{*} | b_j \rangle \, \langle b_j | s_i \rangle .
\label{jankan}
\end{eqnarray}
\begin{figure*}
  \includegraphics[width=0.75\textwidth]{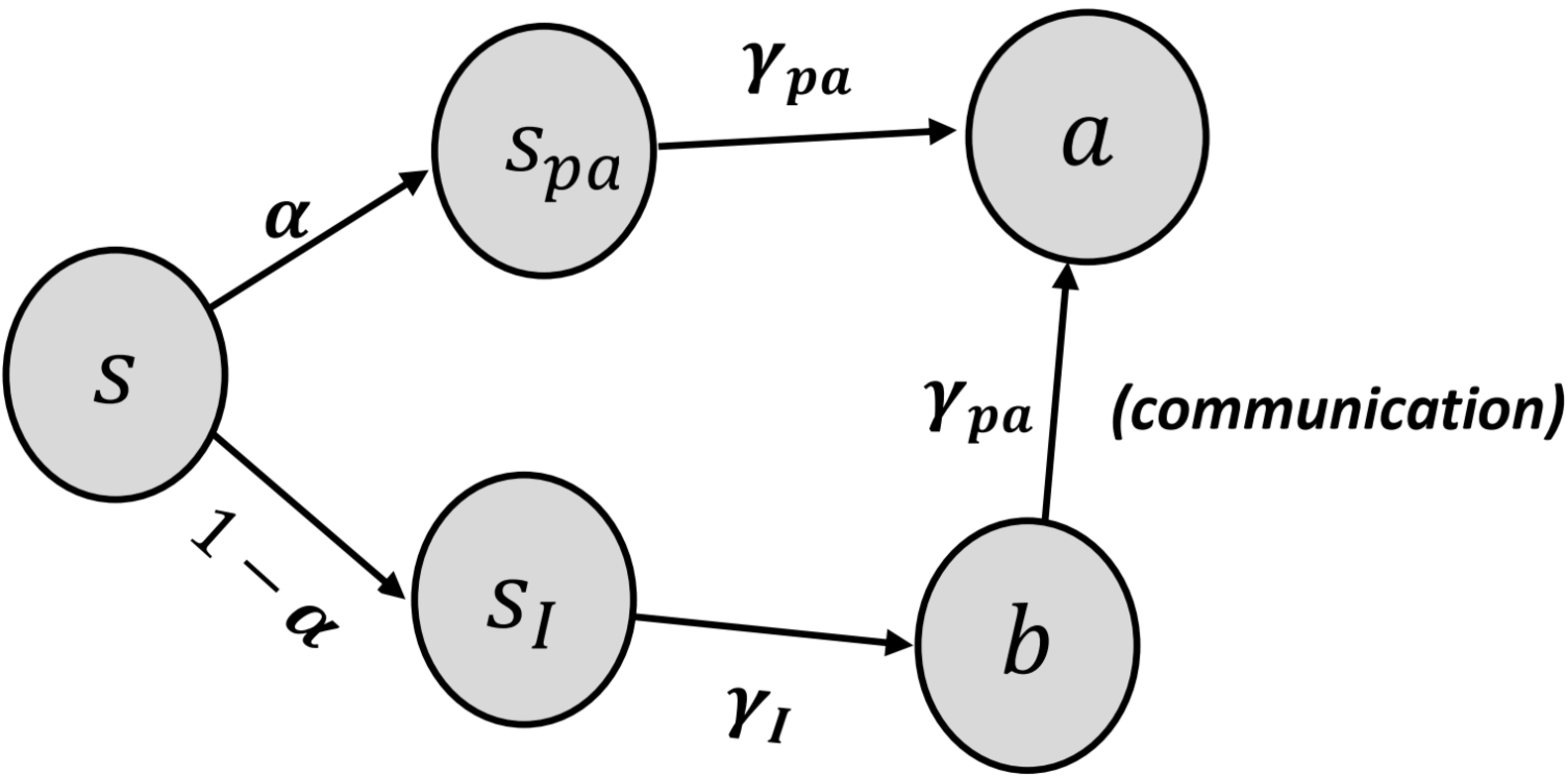}
\caption{A fully joint observable 2-agent network including a given protagonist agent and its interpreter on the basis of Eq.\ref{jankan}.}
\label{serideckheiri}       
\end{figure*}
A network representation for Eq.~\ref{jankan} is illustrated in Fig.~\ref{serideckheiri}. From now on, we will refer to $ r $ as the efficiency (the real part of $ r $ in Eq.\ref{jankan}). 

Equation~\ref{eqktswmtt} is well-defined for our purpose in the current study. Because the operators as $ | b_j \rangle \langle s_j | $ and $ | s_j \rangle \langle b_j | $ for $ b_j \neq s_j  $ cause seemingly nonsensical probabilities in the current study, the former will bring probabilities from which, when a direct world state $ s_j $ is seen, the action takes place for an indirect $ b_j $, and the effect is just the opposite for the latter.\footnote{Operators such $ | c \rangle \langle c | $ in the quantum regime refer to a quantum interference. In this article, however, we are focusing on the classical learning and such quantum issues could be considered for another study}

Suppose that we have derived the asymptotic probabilities as a function of forgetting factors,\footnote{For a specific reward function, see Appendix.~\ref{erageffredg}}
\begin{equation}
{\langle a_k | s_l \rangle}_{max} = {\langle a_k | b_l \rangle}_{max} = 
\begin{cases}
p ( \gamma_{pa} ) , \quad k = l \\
q ( \gamma_{pa} ) , \quad k \neq l 
\end{cases} \\
,{\langle b_k | s_l \rangle}_{max} = 
\begin{cases}
p ( \gamma_{I} ) , \quad k = l \\
q ( \gamma_{I} ) , \quad k\neq l 
\end{cases}
\end{equation}
, where $ \gamma_{pa} $ and $ \gamma_{I} $ stand for the protagonist agent's forgetting factor and interpreter's forgetting factor, respectively. Hence, for a simple fully observable problem with $ a_{s_i}^{*} = a_i $, we have $ r_{max} = p ( \gamma_{pa} ) $, according to Eq.~\ref{eqoriggin} or Eq.~\ref{eqfullyy}.    
Yet, considering our partially observable two-agent model with $|S| = |A| = |B|= N$ and $ N'=N $, one can use Eq.~\ref{jankan} according to $ r_{PO} $, for the asymptotic efficiency in a partially observable environment (versus $ r_{FO} $ preserved for a fully observable one) and write

\begin{multline}
r_{PO} \equiv r_{max}^{pa} (\alpha , \gamma_{pa}, \gamma_{I} ) =  \alpha \, p ( \gamma_{pa} ) + (1- \alpha ) \left[ p ( \gamma_{pa} ) . p ( \gamma_{I} )  + q ( \gamma_{pa} ) . q ( \gamma_{I} ) \right].
\label{eqrmaxxx}
\end{multline}\\[0.1cm]
\begin{equation}
r_{PO} = p ( \gamma_{pa} ) \left\lbrace 1 - (1 - \alpha ) \left[ 1 - p ( \gamma_{I} ) - \frac{q ( \gamma_{I} )}{p ( \gamma_{pa} ) } . q ( \gamma_{I} ) \right] \right\rbrace .
\label{lalbkusab}
\end{equation}
By remembering the efficiency of a given agent in a fully observable environment, $ r_{FO} = p ( \gamma_{pa} ) $, then
\begin{equation}
r_{PO} = \beta \,\, r_{FO} , \qquad 0 \leq \beta \leq 1 \quad \Longrightarrow \quad  r_{PO} \leq r_{FO}.
\label{eq2020}
\end{equation}
We may recall $ \beta $ as a transparency coefficient or $ \mu = 1 - \beta $ as the reduction of transparency. Besides, one can have $ q $ with respect to $ p $ according to the specific form of $ h $-matrix within a given problem; see Eq.~\ref{diwwonehhha}, for instance, for which 
\[ If \quad p + q \, ( N-1 ) = 1, \]
\begin{equation}
\mu = (1 - \alpha ) \, q ( \gamma_{I} ) \left[ (N - 1) - \frac{q ( \gamma_{pa} )}{p ( \gamma_{pa} ) } \right]  , \quad N \geq 2.
\end{equation}\\[0.1cm]  

\begin{figure*}
  \includegraphics[width=0.75\textwidth]{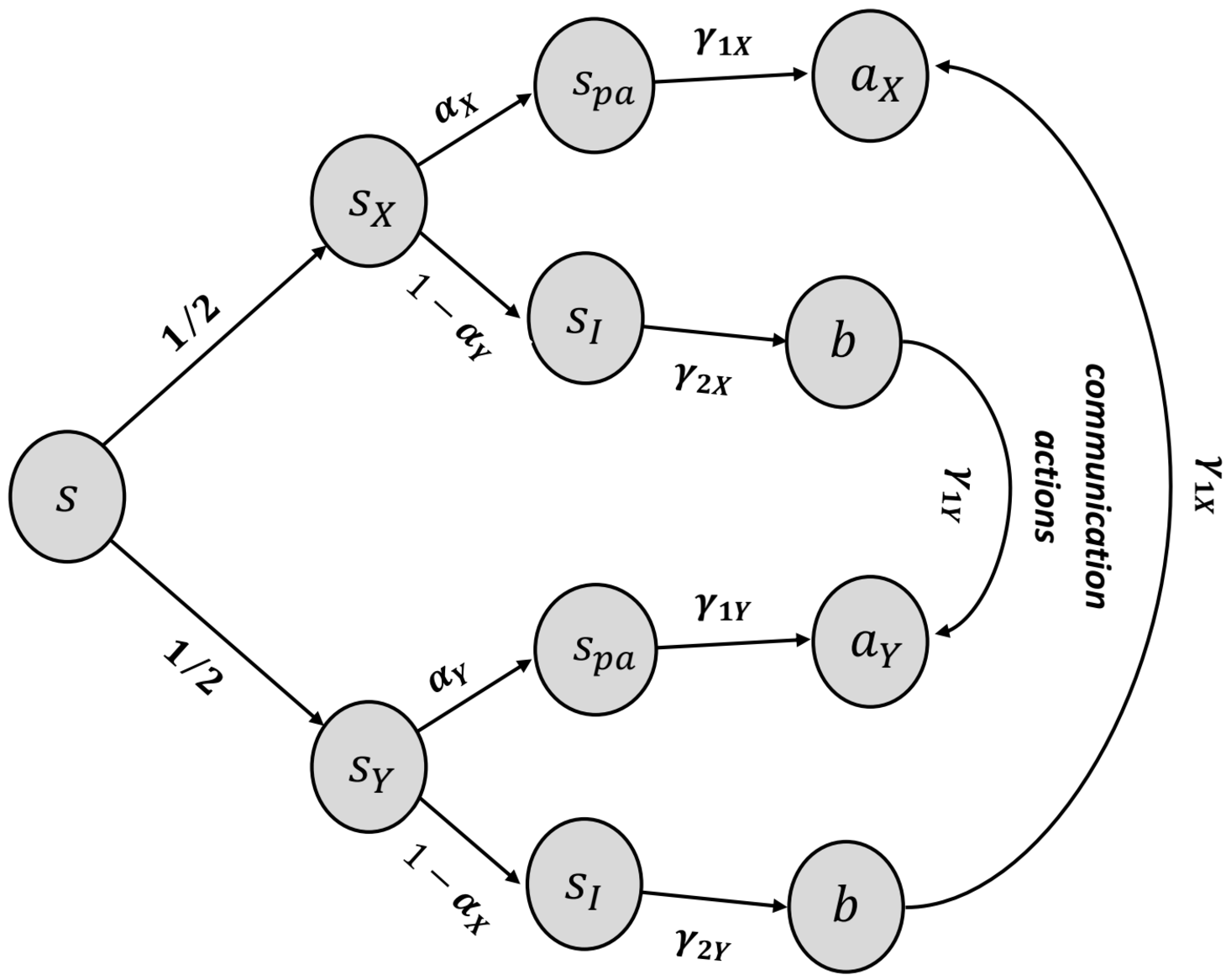}
\caption{A clips network representation including two protagonist-interpreter agents according to Eq.~\ref{gamasamam} and Eq.~\ref{jankan}. Here, there are two blocking and teaching tasks for every agent, namely, $ \gamma_{X} = \gamma_{1X}  + \gamma_{2X} $ and
$ \gamma_{Y} = \gamma_{1Y}  + \gamma_{2Y} $. Furthermore, there could be different observability parameters for every agent ($ \alpha_{X} \neq  \alpha_{Y} $). A relevant situated scenario can be seen in Sec.~\ref{sectionmmultii}}
\label{seridecps}       
\end{figure*}

Our formulation can go beyond to include multi-agent games where an interpreter could be a sort of protagonist agent by itself; a protagonist agent, on the other hand, would be an intelligent interpreter too, so that every agent becomes a simultaneous player-interpreter. Then, some world percepts would be invisible for each agent while another agent can detect them and help its partner to have more efficiency. Assuming a different forgetting factor for every task to be done for a given agent $ X $ in a multi-agent setting containing $ n $ tasks, we could have
\begin{equation}
\gamma_{X} = \sum_{k=1}^n \gamma_{kX} , \quad  0 \leq \gamma_{X} \leq 1.
\label{gamasamam} 
\end{equation}
In such scenarios, the parameter of observability $ \alpha $ can also differ for the two given agents $ X $ and $ Y $, that is $ \alpha_{X} \neq  \alpha_{Y}$. Moreover, it can be realized that the communication is costly unless the forgetting factor of the domain actions is set to be zero. In this particular case (that we recognize as selfishness), communication becomes free. A clips network representation for such a circumstance is illustrated in Fig.~\ref{seridecps} and a relevant example is elaborated in Sec.~\ref{sectionmmultii}.
\subsection{More than two agents} 
\label{kaksobge2}
\begin{figure*}
  \includegraphics[width=0.75\textwidth]{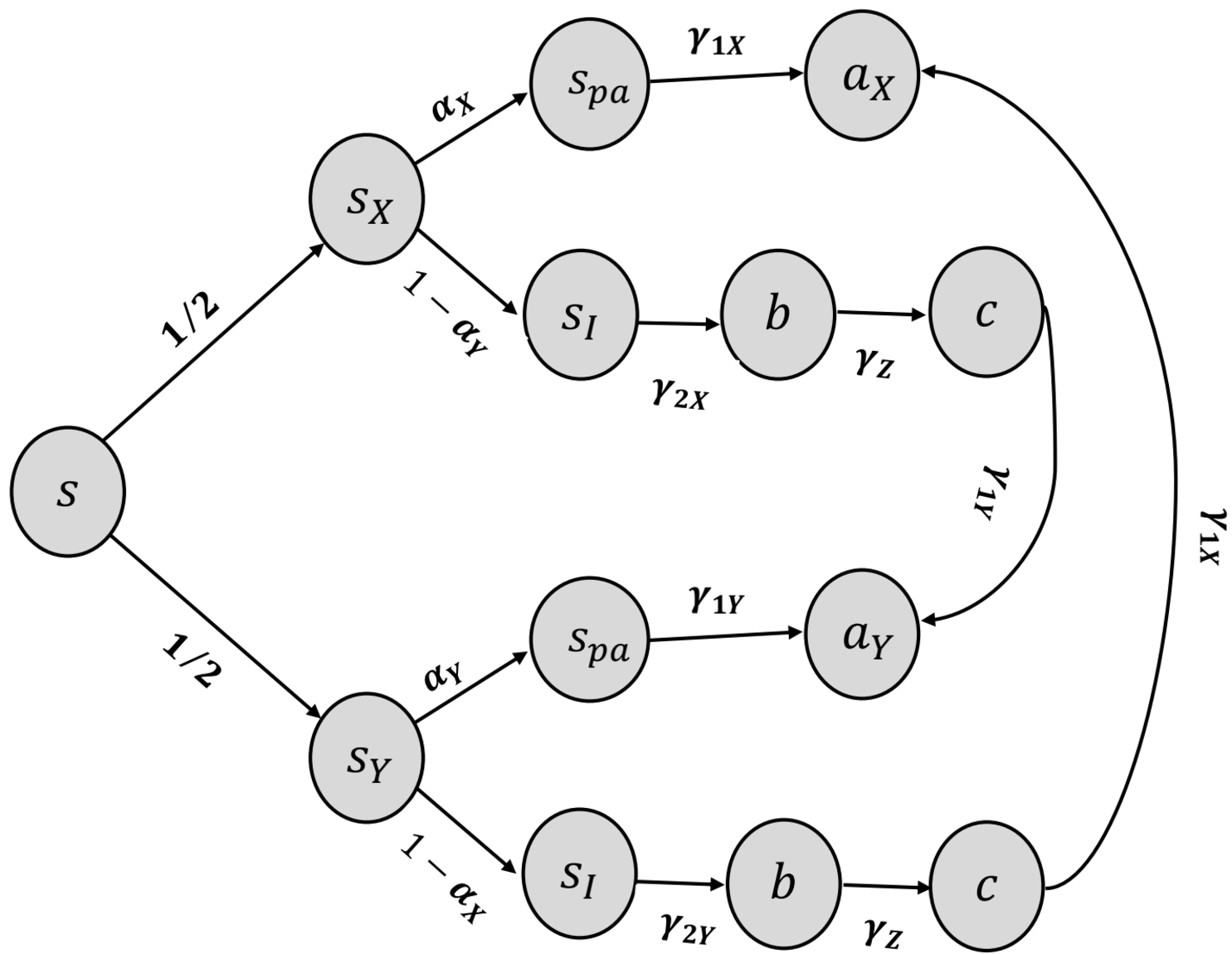}
\caption{A given clips network representation for Eq.~\ref{jankanjfl}, where an additional agent $ Z $ is introduced for a two-folded interpretation task. Here we assume that $ Z $ is an additional interpreter for both $ X $ and $ Y $}
\label{3agsmkdenall3}       
\end{figure*}

For three agents, I can think of two possible extensions; one is as 
 \begin{eqnarray}
efficiency \equiv \frac{\alpha }{N} \sum_{i=1}^N \, \langle a_{s_i}^{*} | {s_i} \rangle \, + \, \frac{1 - \alpha }{N} \sum_{i=1}^N \sum_{j=1}^{N'} \, \sum_{k=1}^{N''} \, \langle a_{s_i}^{*} | c_k \rangle \, \langle c_k | b_j \rangle \, \langle b_j | s_i \rangle 
\label{jankanjfl}
\end{eqnarray} 
, for which a given clips network representation is depicted in Fig.~\ref{3agsmkdenall3}. 
In this form, the agents $ X $ and $ Y $ are both protagonist-interpreter and the agent $ Z $ is an interpreter of the interpreter. In this form, $ Z $ can be an additional interpreter for $ X $ or $ Y $ or both of them. 

\begin{figure*}
  \includegraphics[width=0.6\textwidth]{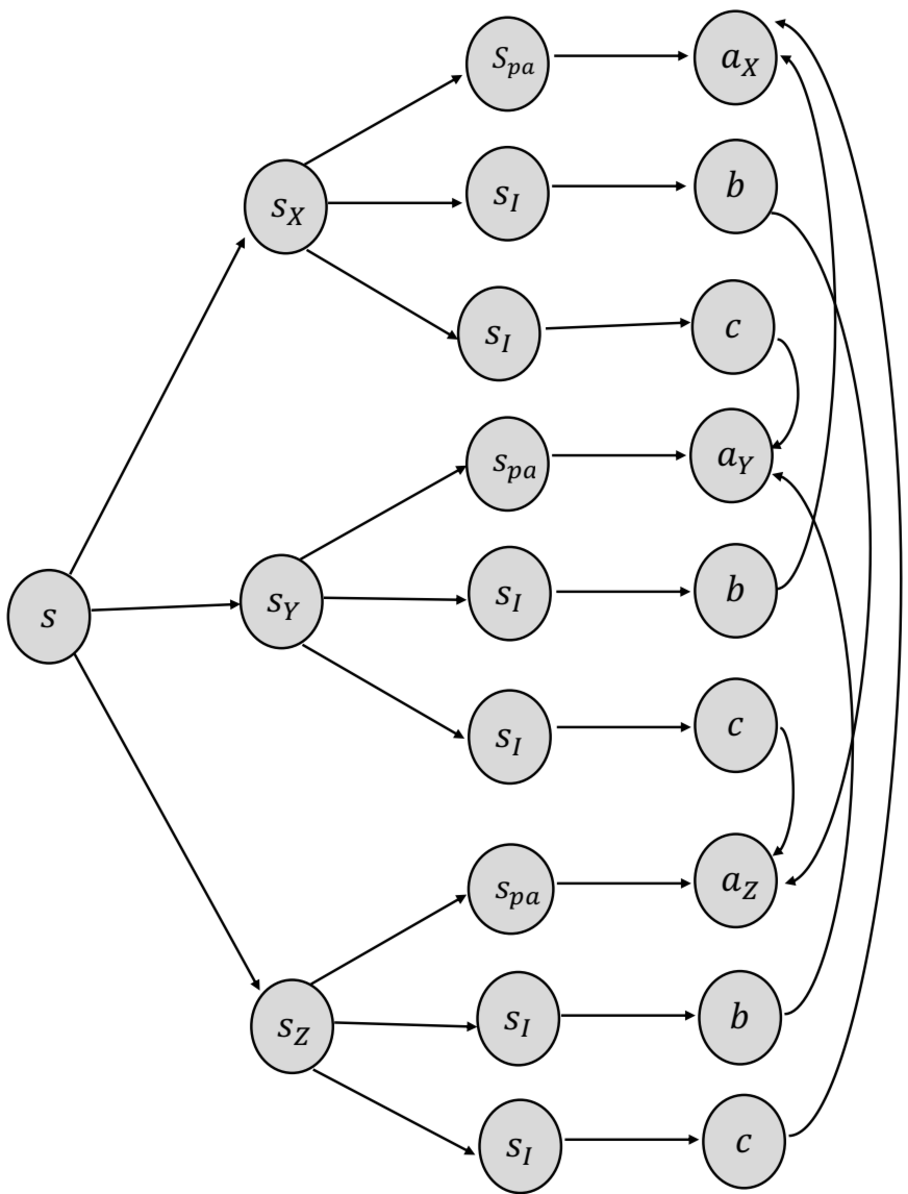}
\caption{A given clips network representation for Eq.~\ref{janflahspkan}. Here, there are three protagonist-interpreter agents and the world states remain jointly fully observable}
\label{serid3agecps}       
\end{figure*}

Another extension could be having three protagonist-interpreter agents; let $ \alpha \rightarrow \alpha_1 $, $ 1 - \alpha = \alpha_2 + \alpha_3 $, so that the related efficiency will have three terms as
\begin{eqnarray}
\frac{\alpha_1 }{N} \sum_{i=1}^N \, \langle a_{s_i}^{*} | {s_i} \rangle \, + \, \frac{ \alpha_2 }{N} \sum_{i=1}^N \sum_{j=1}^{N'} \, \langle a_{s_i}^{*} | b_j \rangle \, \langle b_j | s_i \rangle + \, \frac{ \alpha_3 }{N} \sum_{i=1}^N \sum_{k=1}^{N''} \, \langle a_{s_i}^{*} | c_k \rangle \, \langle c_k | s_i \rangle  .
\label{janflahspkan}
\end{eqnarray}
A given clips network representation for Eq.~\ref{janflahspkan} is illustrated in Fig.~\ref{serid3agecps}. 

The generalization of this method to N agents is then straightforward, such that it includes all of the possible terms of the efficiency as 
\begin{eqnarray} 
& \sum \langle a_{s_i}^{*} | {s_i} \rangle \notag + \\ 
 & \sum \langle a_{s_i}^{*} | b_j \rangle \, \langle b_j | s_i \rangle + \sum \langle a_{s_i}^{*} | c_j \rangle \, \langle c_j | s_i \rangle + \cdots \notag \\
& + \sum \langle a_{s_i}^{*} | c_k \rangle \, \langle c_k | b_j \rangle \, \langle b_j | s_i \rangle +  \cdots \notag \\
& + \sum \langle a_{s_i}^{*} | d_l \rangle \, \langle d_l | c_k \rangle \, \langle c_k | b_j \rangle \, \langle b_j | s_i \rangle + \cdots \notag \\ 
& + \cdots \notag \\
&  \:  \vdots
\end{eqnarray}

In our introduced framework, communication actions are costly informative messages which are not noise-free unless $ \gamma_{I} = 0 $. One other option could be when an agent can decide to communicate or not. In other words, some teaching agents might decide not to send even an empty message from which a protagonist agent may make use by performing at least a random action. 

Last, depending on the objective of learning \cite{Shoham}, a generalized belief state is needed built upon the current world state, the observability parameter $ \alpha $, and the distribution of the forgetting factors, etc.\footnote{For example, if the game is known or unknown or given the purpose of a game, a coalitional strategy might be useful}A suggestion for building such a more complicated belief state could be the use of a meta-learning approach \cite{Briegel3} considering a multi-agent setting.
     
\section{Specific examples of a 2-agent invasion toy problem}
The original formulation of PS has been introduced using a toy problem called invasion game, as elaborated in \cite{Briegel1,Briegel2}. To add partial observability employing another agent to the original projective simulation, we can assume an interpreter added to the standard invasion problem. In the basic form of the invasion game, an attacker (A) sends some precepts $s \in \{ \Leftarrow , \Rightarrow \}$ (in case $ N=2 $) where a defender (D) perceives and learns them by taking action $a\in \{ - , + \}$ on a percept and getting a reward ($\lambda$).  
\begin{equation}
 \{ s_1 , s_2 \} = \{ \Leftarrow , \Rightarrow \}, \qquad
\{ - , + \} = \{ a_1 , a_2 \} ,
\end{equation}
where we can consider $ | s_i \rangle = \{ | \Leftarrow  \rangle , | \Rightarrow \rangle \} $ and $ | a_i \rangle = \{ | -  \rangle , | + \rangle \} $ for $ N=2 $ in an invasion game. 

While the theoretical asymptotic efficiency for a small amount of $ \gamma $ has been derived for a fully observable one-defender invasion game \cite{Briegel1,Briegel2}, that derivation cannot be used for an arbitrary choice of $ \gamma $, even in the simple form of one agent toy problem. The asymptotic efficiency related to a given agent ''$ D $'', that is $r_{max}^D = r_{max}^D ( \alpha , \gamma_{D}, \gamma_{I}) $, occurs with respect to $ \lim_{t \to \infty} h^{(t)} (s,a) $, based on the adaptation rule of Eq.~\ref{eqtrans}. This adaptation is changed stochastically, however, in a big enough time, $ t \to \infty $; the averaged efficiencies on a large number of agents ($ m $) reach a certain asymptote for every forgetting factor $ \gamma $ (see Appendix.~\ref{averagefequa}). Then one can assume that there is an effective reward function, $ \lambda_{eff} $, for every $ \gamma $, so that it becomes the averaged reward obtained on a large number of agents $ m $.    
\begin{equation}
{\left\langle \sum_l \delta (c_i , c_{k_l} ) \delta (c_j , c_{m_l}) {\lambda}^{(t+1)} \right\rangle }_{ m \to \infty} \longrightarrow \quad \lambda_{eff}^{(t+1)} (c_i , c_j , \gamma ) ,
\end{equation}
thus,
\begin{equation}
h^{(t+1)} (c_i , c_j ) = h^{(t)} (c_i , c_j ) - \gamma ( h^{(t)} (c_i , c_j ) -1 ) + \lambda_{eff}^{(t+1)} (c_i , c_j , \gamma ) .
\label{effectivee}
\end{equation}\\[0.1cm]
Furthermore, a constant reward in every time in Eq.\ref{eqtrans}, namely, 
\begin{equation}
{\lambda}^{(t+1)} (c_i , c_j ) = \lambda (c_i , c_j ),
\label{conditionddq}
\end{equation}
leads to
\begin{equation}
h_{max} = \lim_{t \to \infty} h^{(t)} (c_i , c_j ) \, = \, \frac{\lambda_{eff} (c_i , c_j , \gamma ) }{\gamma} + 1. 
\end{equation}  
Then,
\begin{equation}
h_{max} =
\begin{cases}
\frac{\lambda_{eff}^{reward}}{\gamma} + 1 & \textit{for a rewarded pair of $ (c_i , c_j) $} \\
1 & \textit{for a non-rewarded pair of $ (c_i , c_j) $}
\end{cases}
\end{equation}
Due to the fact that $ \lambda_{eff}^{non-reward} = 0 $, we rename $ \lambda_{eff}^{reward} = \lambda ( \gamma ) $ from now on. Given $ N $ different percept-actions but just one desirable action $ a_{s_i}^{*} $ for each percept $ s_i $, we have
\begin{equation}
p + q \, ( N-1 ) = 1,
\label{diwwonehhha}
\end{equation}
and 
\begin{equation}
p_{N} ( \gamma ) = \frac{\lambda_{N} (\gamma) + \gamma}{\lambda_{N} (\gamma) + N \gamma} \quad , \quad q_{N} ( \gamma ) = \frac{ \gamma}{\lambda_{N} (\gamma) + N \gamma} , \qquad p(0) = 1, \qquad q(0) = 0. 
\label{incsioo}
\end{equation}

Equations~\ref{incsioo} are the evaluation of the asymptotic probabilities in our partially observable multi-agent model of problems for which Eq.~\ref{conditionddq} is satisfied. Otherwise, we may use the specific properties of a given scenario to have $ p $ and $ q $.

\begin{figure*}
  \includegraphics[width=0.75\textwidth]{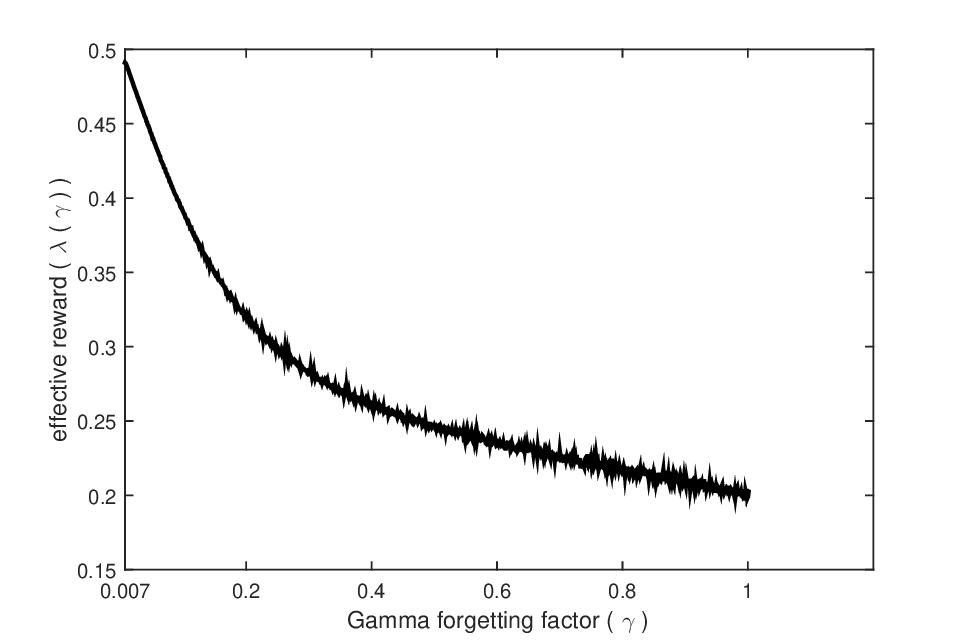}
\caption{Effective reward function ($ \lambda_{eff}^{reward} $) of a rewarded $ (c_i, c_j ) $ for a fully observable invasion toy problem with $ N=2 $ with regard to $ \lambda^{t+1} =1 $ for two rewarded pairs of $ (s,a) =\{ (\Rightarrow , +), (\Leftarrow , -) \} $, if traversed at the last time $ t $, and $ \lambda^{t+1} =0 $ for other edges. The results are averaged on $ m=10000 $ agents in the time $ t=1000 $ }
\label{effrewardx}       
\end{figure*}

In the following, we consider an invasion with $ N=2 $. Figure~\ref{effrewardx} shows the effective reward of the rewarded pairs of $(c_i, c_j )$ for a fully observable one-agent invasion with $ \lambda^{t+1} (s, a_{s}^{*} ) =1 $.
For instance, 
\begin{equation}
p_2 (\gamma \rightarrow 1) \longrightarrow \frac{1.2}{2.2}, \quad q_2 (\gamma \rightarrow 1) \longrightarrow \frac{1}{2.2}.  
\end{equation}
\subsection{An example of an absolute partially observable environment according to Equation \ref{blockeff}}
\begin{figure*}[h]
  \includegraphics[width=0.75\textwidth]{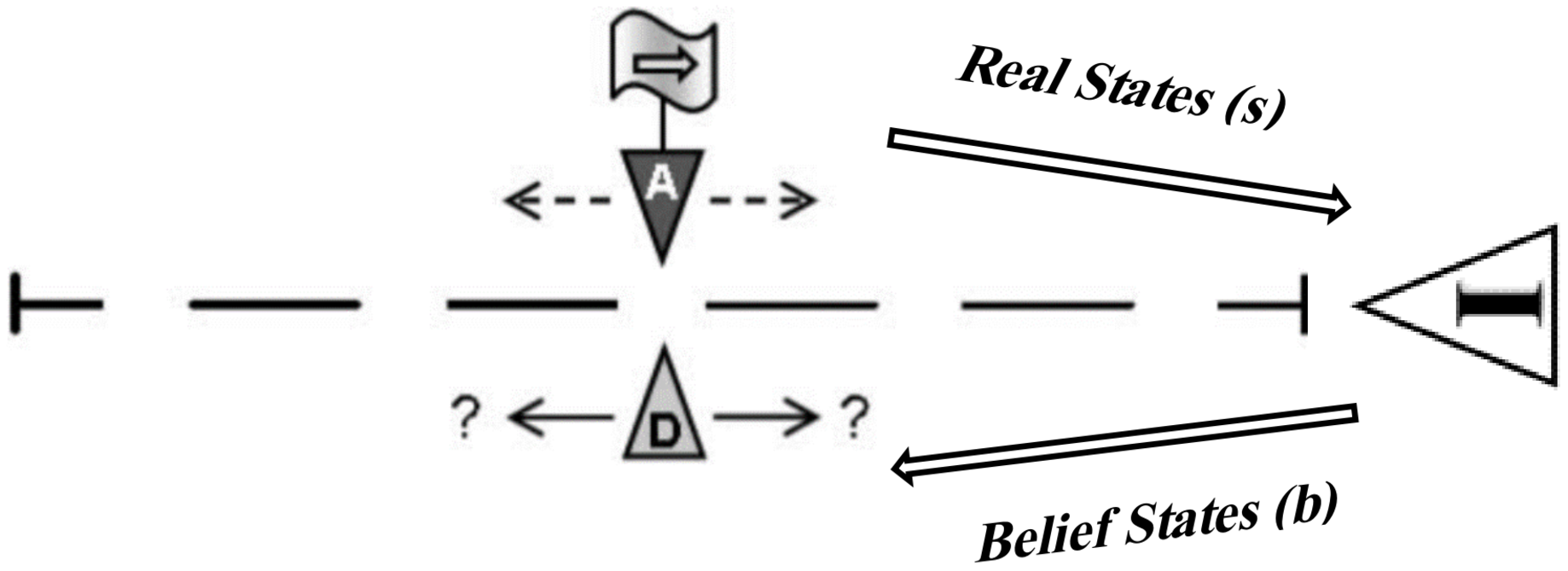}
\caption{A model of the partially observable invasion game. Adapted and modified from \cite{Briegel1}. In addition to an attacker (A) and a defender (D), we have an interpreter (I) here. While ''D'' can not detect the attacker's percepts, ''I'' observes these signs, learns them, and sends them as some new percepts to the defender.}
\label{interpreter:1}       
\end{figure*}

Fully observable environments have been considered widely under different conditions. Here, we start from Eq.~\ref{blockeff}, which describes an environment in which all of the percepts are invisible to the protagonist agent. In this case, the belief percepts are produced by an interpreter (I), such as what is depicted in Fig.~\ref{interpreter:1}, which comes indirectly to the defender (D). To be nontrivial, the interpreter is not a kind of a simple mirror or a polarizer. Instead, it is another intelligent agent that learns the percepts $ s_i $ by itself and sends what it is learning to the defender as a new percept $ b_i $, where the defender percieves them and takes actions $ a_i $ on them. As it is assumed in the original paper \cite{Briegel1}, the defender always reaches sooner to the next door than the attacker does (with or without a stamp from the interpreter). The relevant reward function definition can be found in Appendix.~\ref{erageffredg}.   

\begin{figure*}[h]
\centering
\subfigure[]{
\includegraphics[width=5.6cm, height=4.7cm]{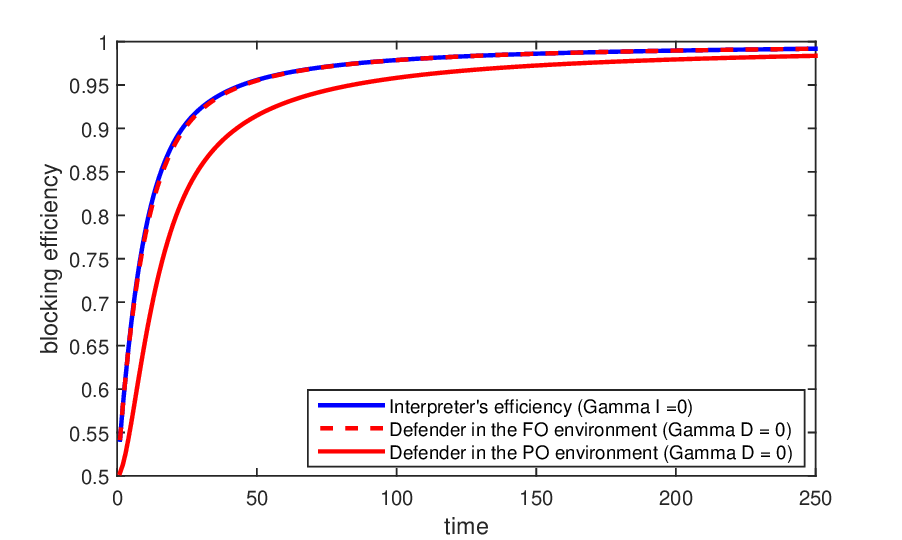}
\label{n_027000}
}
\subfigure[]{
\includegraphics[width=5.6cm, height=4.7cm]{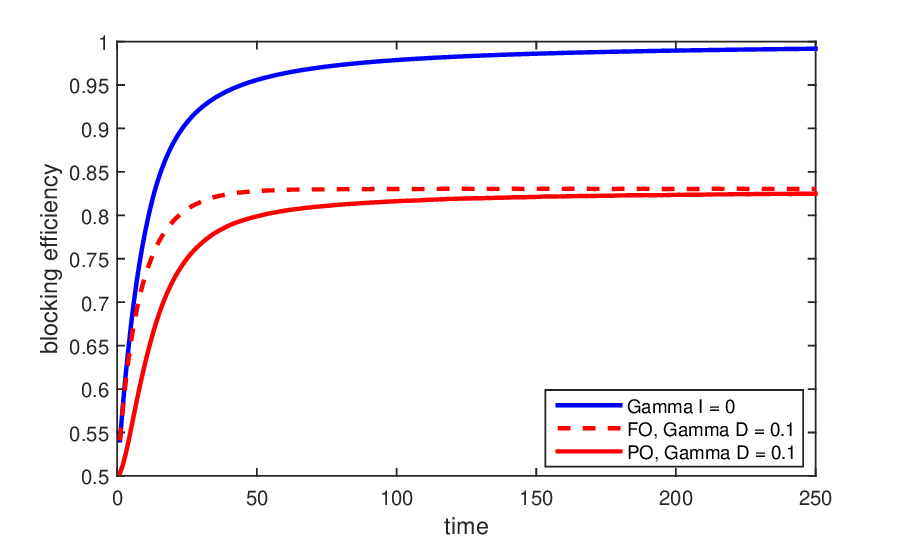}
\label{n_227000}
}
\subfigure[]{
\includegraphics[width=5.6cm, height=4.7cm]{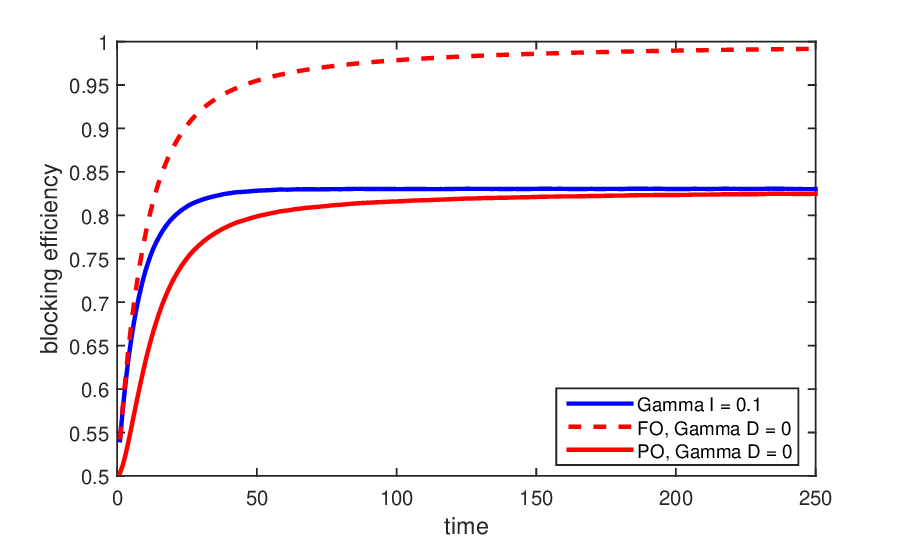}
\label{n_127000}
}
\subfigure[]{
\includegraphics[width=5.6cm, height=4.7cm]{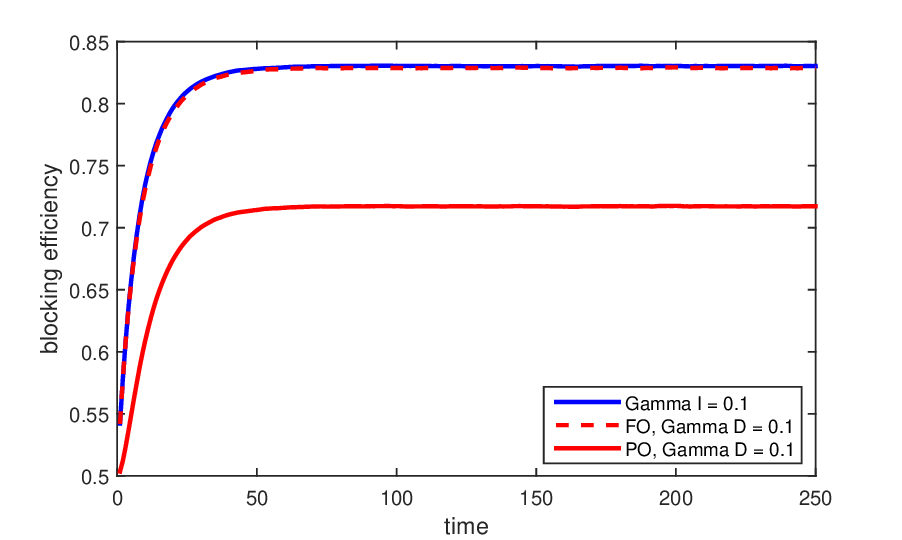}
\label{n_1027000}
}
\caption{Comparison between blocking efficiencies of a defender in a partially observable (PO) invasion game (red solid curves) and a fully observable (FO) invasion game (dashed) with regard to different forgetting factors for both defender (D) and interpreter (I). Results for the interpreter (blue curves) are illustrated using Eq.\ref{eqfullyy}, and the defender's results are illustrated using  the blocking efficiency, $ r^{(t)} $, of (Eq.~\ref{blockeff}). All of the curves are averaged over 10000 agents}
\label{gasbulbdata}
\end{figure*}

Figures \ref{n_027000} and \ref{n_227000} show a reduction in the speed of learning (or learning time) in a partially observable environment (red solid curves) in comparison with a fully observable one (dashed curves) when the interpreter's forgetting factor is zero. In contrast, there is a reduction in the efficiency of the defender, as illustrated in \ref{n_127000} and \ref{n_1027000}, when the interpreter's forgetting factor is more than zero. While the reduction in the learning time is due to the fact that the defender learning must wait for the interpreter learning, the non-vanishing decrease in the efficiency is owing to the portion of permanent partially observability arising from the interpreter's forgetting factor.\footnote{In the original works, the authors have plotted their figures for action blocking, whereas mine have been planned in respect to the probability of action blocking using the fact that action blocking averaged on an infinite number of actions would ultimately be equal to the probability of doing an action; for more details, see Appendix \ref{averagefequa}.  }   

\begin{figure*}[h]
\includegraphics[width=5.6cm, height=4.7cm]{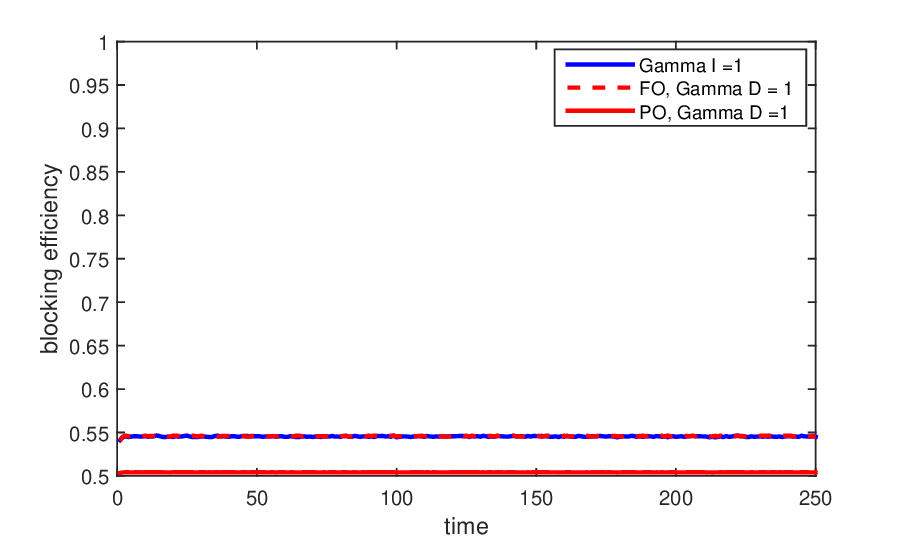}
\caption{Comparison of the blocking efficiency of a defender in an absolute partially observable invasion (red solid line) and in the fully observable invasion (dashed) based on $ \gamma = 1 $ for both defender (D) and interpreter (I)}
\label{gasbu}
\end{figure*}

Figure \ref{gasbu} illustrates the same property in Fig.~\ref{n_1027000} for the extreme dissipation factor of $ \gamma = 1 $. 
It is worth noting that the multiplied probabilities of Eq.\ref{blockeff} reduce the maximum blocking efficiency of a partially observable environment, in comparison with a fully observable one, for every $ \gamma > 0 $, even in the case of the minimum efficiency of $ \gamma = 1 $. This is because the term including $ \gamma $ in the adaptation rule of Eq.~\ref{eqtrans} refers to forgetting what the agent learned in the previous states, but not the current state. Therefore, this adaptation rule leaves something (even small) more than nothing ($ r_{max}  > 0.5 $) for $ \gamma = 1 $ that can be reduced in a partially observable environment. 

\subsection{An example of a general fully-partially observable environment according to Equation \ref{jankan}}
\label{secsecdfga}
In a more general scenario, there are both partially and fully observable percepts in the environment, where an agent is expected to act. A physics correspondence for such a situation might be where two kinds of ray lights are coming to a given (learning!) polarizer in the condition that the $ \alpha $ portion of the rays has already been polarized in the same direction of the polarizer.  
In our invasion example, one can add some portion ($ \alpha $) of fully observability to the problem. It means that the defender can see an $ \alpha $ portion of its percepts directly coming from the attacker and $ 1 - \alpha $ portion coming indirectly from the interpreter. As a tangible example of the situation, one can suppose that the attacker's signs are being sent in two different colors from which one of these colors are invisible for a given defender. 
\begin{figure*}[h]
  \includegraphics[width=0.75\textwidth]{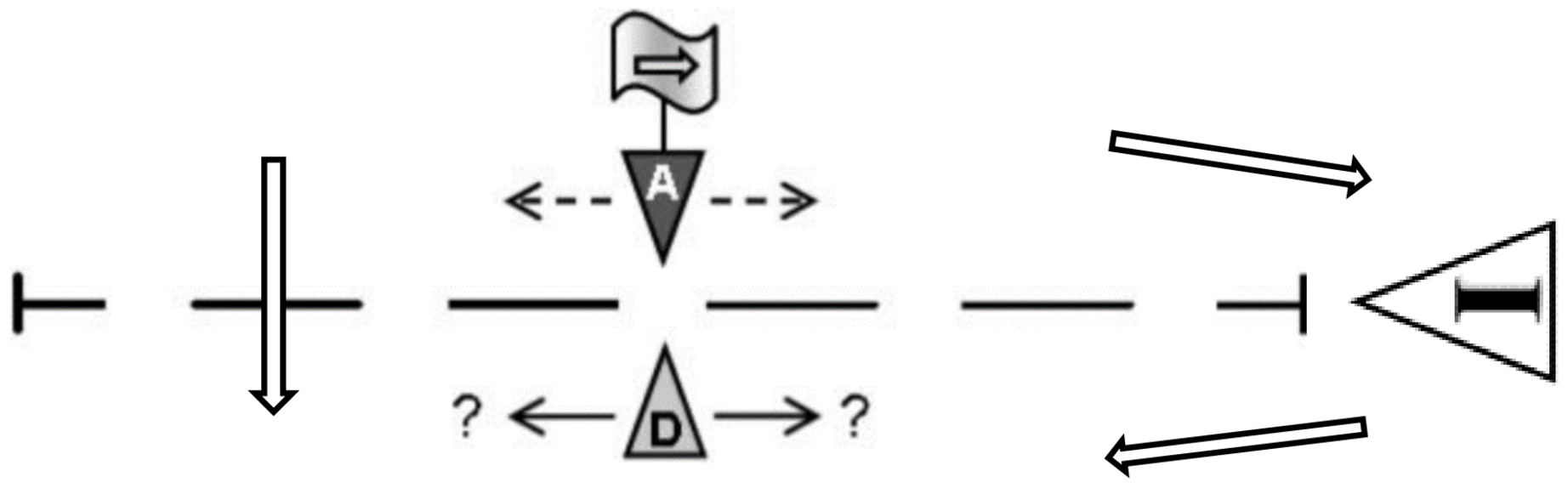}
\caption{A model of the fully-partially observable invasion game, where, in addition to the features of Fig.~\ref{interpreter:1}, a portion of percepts is fully observable for the defender and comes directly from the attacker}
\label{interpreter:2}       
\end{figure*}

\begin{figure*}[h]
\centering
\subfigure[]{
\includegraphics[width=5.6cm, height=4.5cm]{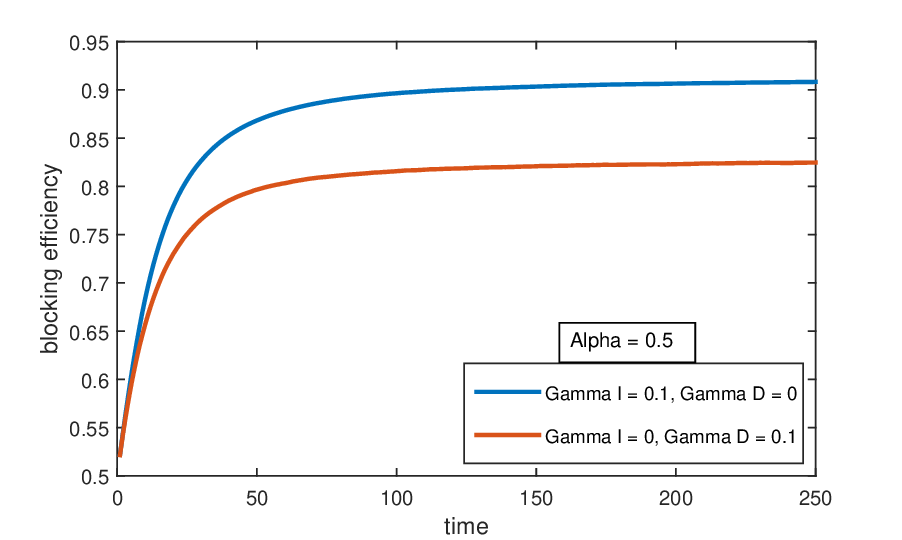}
\label{pilot1}
}
\subfigure[]{
\includegraphics[width=5.6cm, height=4.5cm]{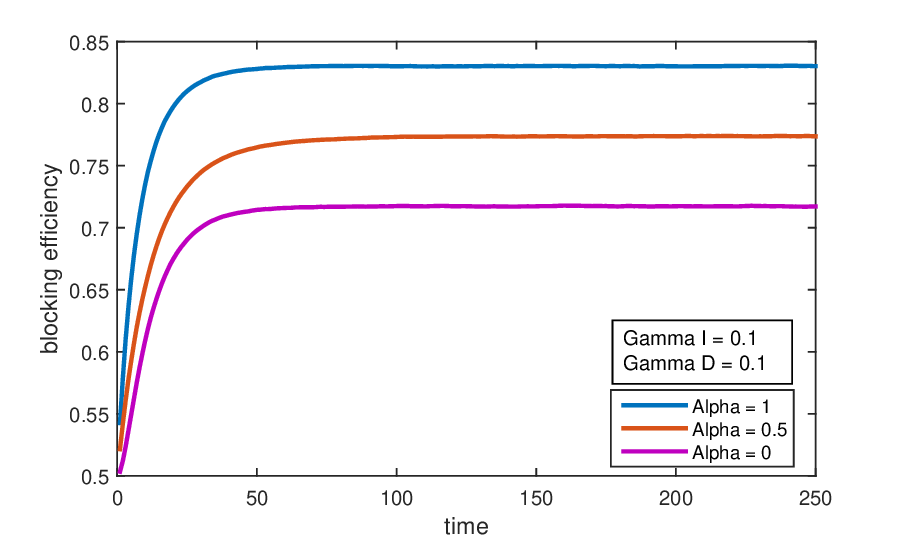}
\label{kesh2}
}
\caption{Blocking efficiency of a defender in the general scenario of a fully-partially observable environment of Fig.\ref{interpreter:2}. \ref{pilot1}: with a certain portion of direct percepts ($ \alpha $), the direct dissipating factor (Gamma D) reduces the efficiency much more than the indirect (Gamma I), \ref{kesh2}: The reduction in the amount of the fully observable percepts ($ \alpha $) can reduce the amount of efficiency}
\label{pilotpkesh}
\end{figure*}

Figures.~\ref{gasbulbdata} and \ref{gasbu} are specific examples of the current scenario with the observability parameter of $ \alpha =0 $. A comparison between the effect of the defender's forgetting factor ($ \gamma_{D}$) and the interpreter's forgetting factor ($ \gamma_{D}$) is illustrated in Fig.~\ref{pilot1}; that is, the former (Gamma D) dominates the latter (Gamma I) in the amount of efficiency. Moreover, the effect of the reduction of observability $ \alpha $ in the amount of efficiency is depicted in Fig.~\ref{kesh2}. What can apparently be seen from these two, Fig.~\ref{pilot1} and Fig.~\ref{kesh2}, is that the contribution of the defender itself to its efficiency is more important than that of the interpreter. To be more clear, in the following, we focus on the maximum blocking efficiency (asymptotic efficiency) containing both Gamma factors ($ \gamma_{D}$ and $\gamma_{I} $).        
\begin{figure*}[h]
\centering
\subfigure[]{
\includegraphics[width=5.6cm, height=4.7cm]{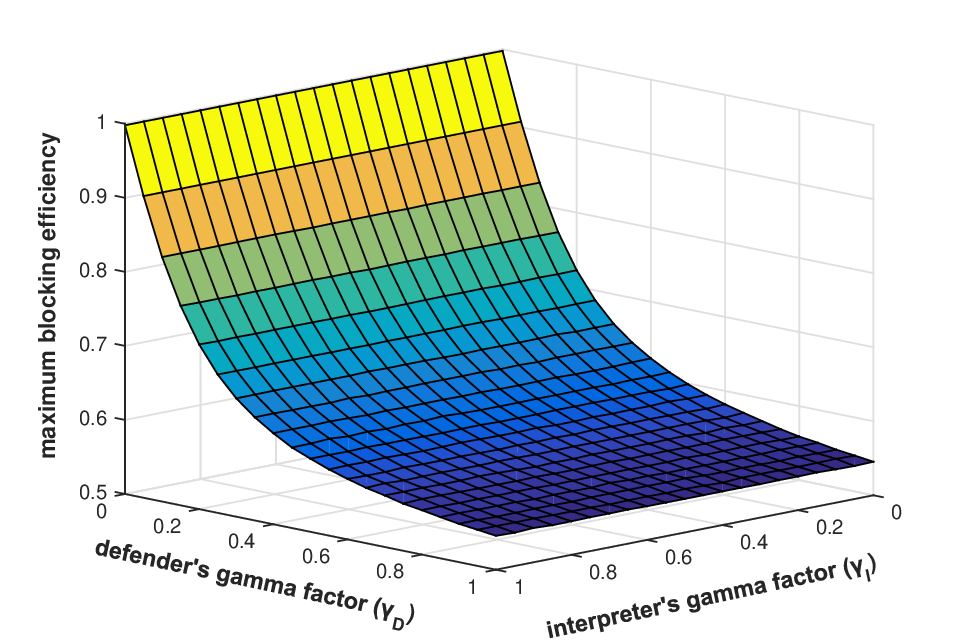}
\label{3d_alpha_1}
}
\subfigure[]{
\includegraphics[width=5.6cm, height=4.7cm]{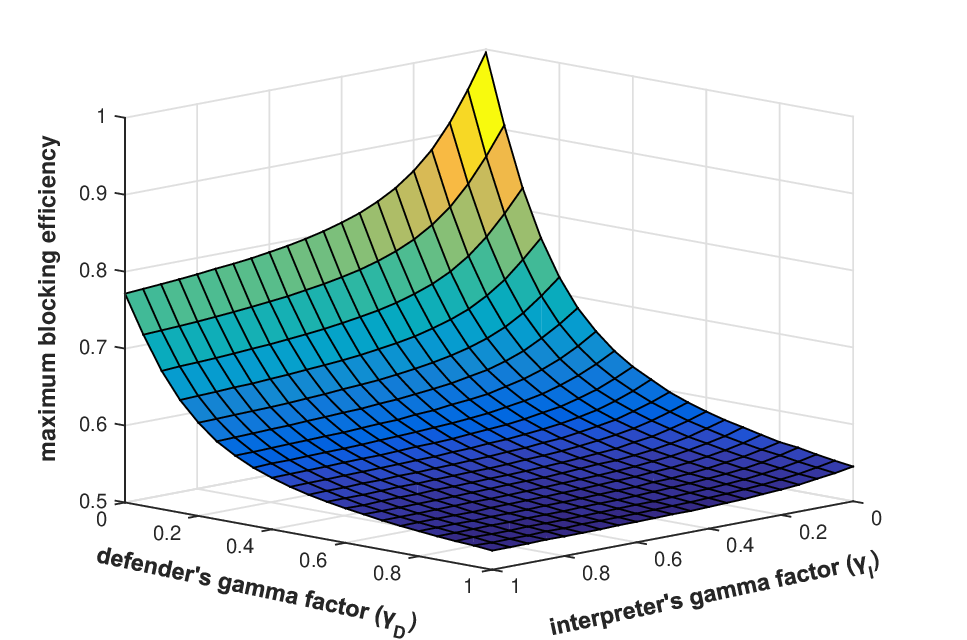}
\label{3d_1_alpha_5}
}
\subfigure[]{
\includegraphics[width=5.6cm, height=4.7cm]{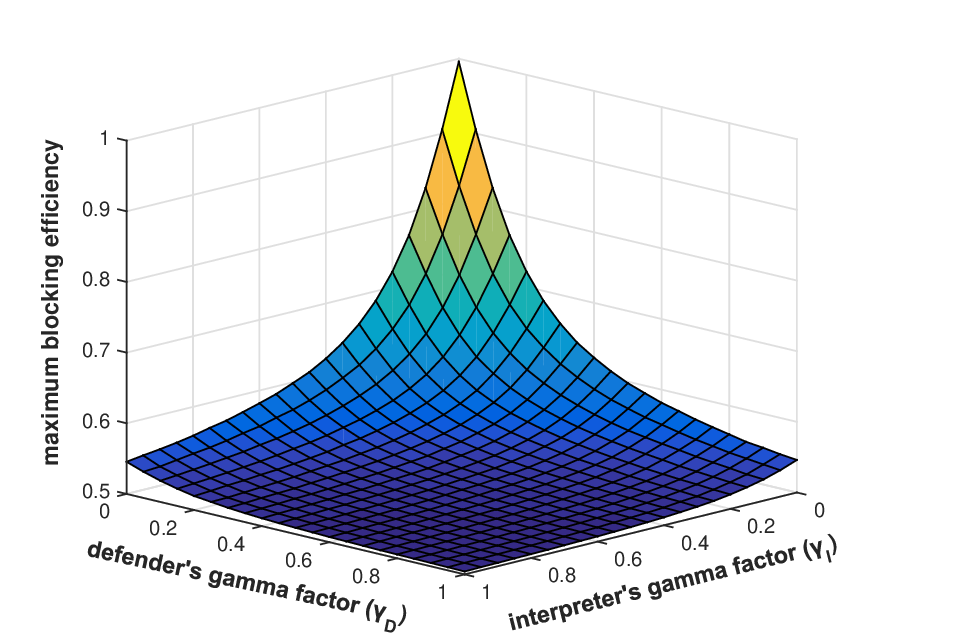}
\label{3d_alpha_0}
}
\caption{Maximum blocking efficiency (asymptotic efficiency) in some different $ \alpha $. Figure~\ref{3d_alpha_1} belongs to $ \alpha =1 $, whereas Figs.~\ref{3d_1_alpha_5} and \ref{3d_alpha_0} are depicted for $ \alpha =0.5 $ and $ \alpha =0 $, respectively. The results averaged over 10000 agents}
\label{3d_alpha_1_0_5}
\end{figure*}
Figure~\ref{3d_alpha_1_0_5} shows that, while the blocking efficiency in the plane of $ \gamma_{D} = constant $ is altered drastically with changing $\alpha$, the plane of $ \gamma_{I} = constant $ has a few changes (respecting $ \alpha $ alterations) in its amounts. It is because the reduction of the observability $ \alpha $ means an increase in the contribution of the interpreter and thus, its dissipation factor $ \gamma_{I} $ becomes more important. Therefore, as it might be expected, we can see that while in $ \alpha =1 $ (Fig.~\ref{3d_alpha_1}), the interpreter's forgetting factor has no role in the amount of maximum efficiency due to the fact that the environment is fully observable in this case, it will be as influential as the defender's gamma factor in $ \alpha =0 $ (the absolute partially observable environment). Therefore, because of the relevant game-theoretical considerations, in the next two asymptotic figures, I will focus just on the plane of $ \gamma_{D} = constant $ to show more details.

\begin{figure*}
  \includegraphics[width=0.75\textwidth]{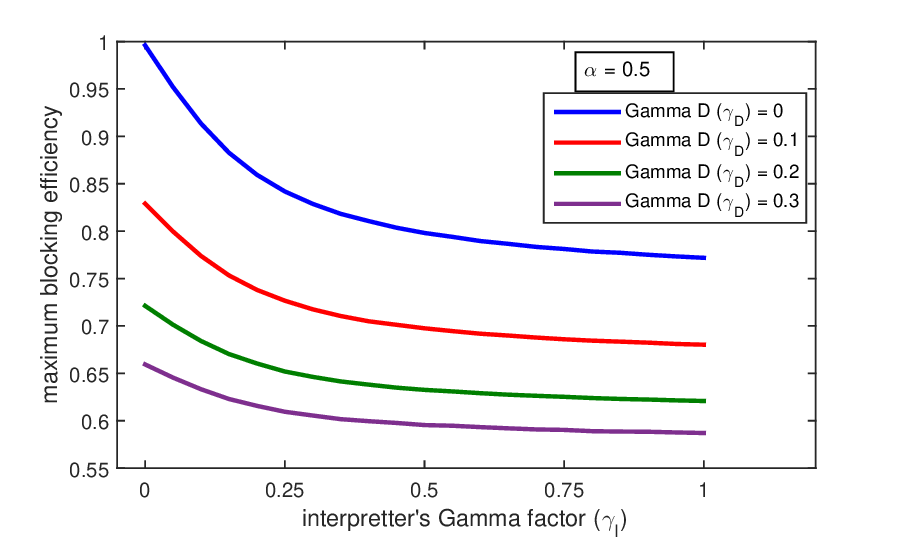}
\caption{Comparison between different maximum blocking efficiencies for a given $ \alpha = 0.5 $ in the slices of $ \gamma_{D} = const $}
\label{alpanim}       
\end{figure*}

\begin{figure*}
  \includegraphics[width=0.75\textwidth]{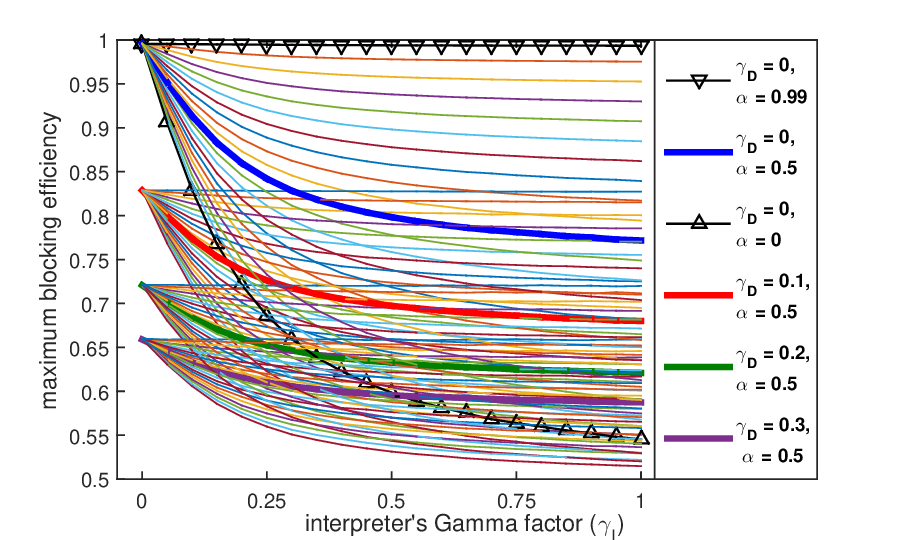}
\caption{A scattering happens in the lines of the maximum blocking efficiencies for $\gamma_{D} = const $ in different $ \alpha $ factors. This scattering causes some intersections in the number of maximum blocking efficiencies which cancel the permanent domination of a smaller $\gamma_{D} = const $ therein}
\label{alpamore}       
\end{figure*}

Eventually, Fig.~\ref{alpanim} compares the maximum blocking efficiencies of the defender in some planes of $ \gamma_{D} = constant $ in a certain amount of $\alpha = 0.5$. Obviously, there is no intersection between the lines in Fig.~\ref{alpanim}, which refers to the domination of a smaller $ \gamma_{D} $ for a given $\alpha$.\footnote{Although Fig.~\ref{alpanim} is depicted just for $ \alpha = 0.5 $, this consequence is true for every $ \alpha = constant $ that can be tracked in Fig.~\ref{alpamore}.} Alternatively, when we also introduce some different $ \alpha $ factor in the planes of $ \gamma_{D} = const $, there is a bunch of plots (or a scattering) for every $ \gamma_{D} = const $; we can see some intersection between the lines of the asymptotic efficiencies thereafter. As a result, in a certain observability ($ \alpha = const $), the line of maximum blocking efficiency dominates for a smaller $ \gamma_{D} = const $; however, the ultimate domination of a smaller $ \gamma_{D} $ can be annihilated by the changes in the amount of the observability of the environment.  
In the next section, we use this consequence to further discuss related a 2-defender-interpreter game. 
\subsection{A given 2-defender-interpreter invasion game}
\label{sectionmmultii}
\begin{figure*}[h]
  \includegraphics[width=0.75\textwidth]{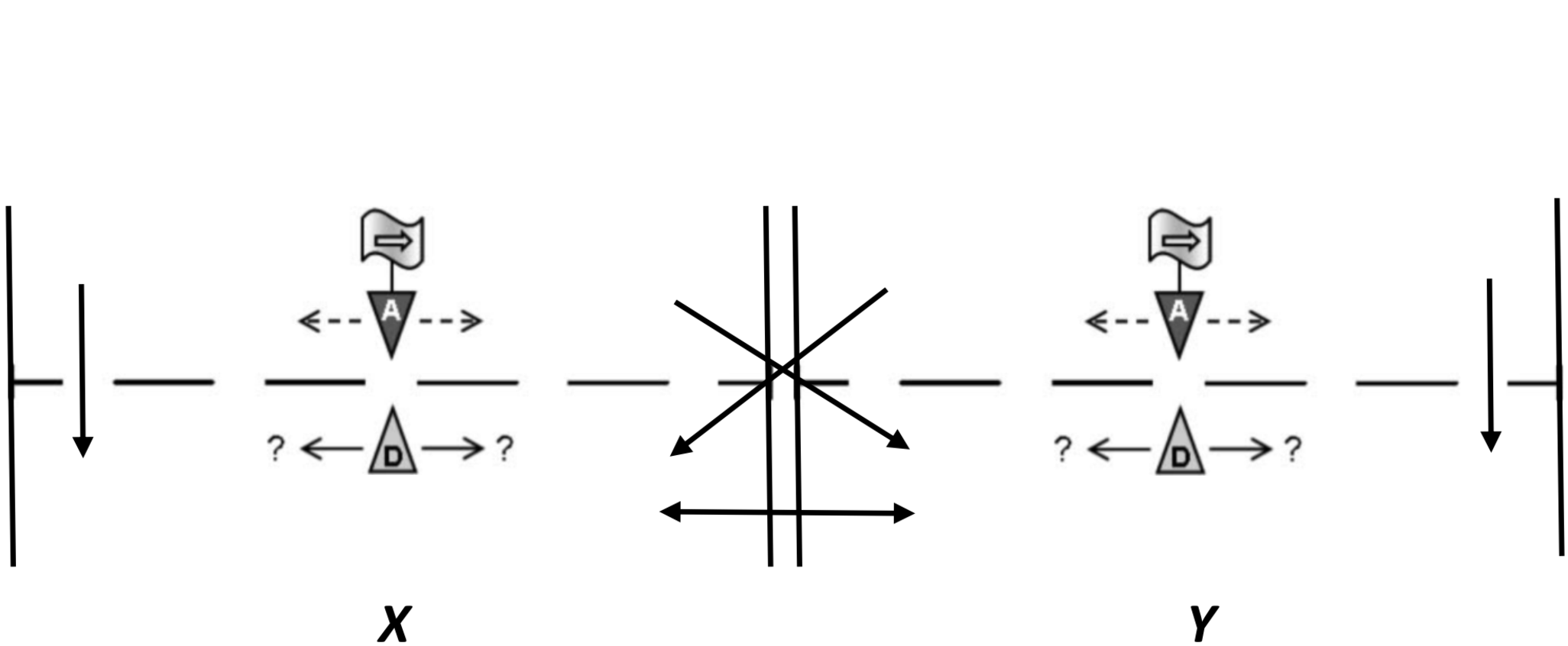}
\caption{A model of a two defender-interpreter invasion game. The scenario is the same as Fig.~\ref{interpreter:2} for every agent. To prevent any interfering percepts, one can consider each agent as a defender (or interpreter) at any time or any round. A specefic clips network representation has been illustrated in Fig.\ref{seridecps}}
\label{fig:1}       
\end{figure*}

In this section, we consider an invasion including two agents $ X $ and $ Y $, as illustrated in Fig.~\ref{fig:1}; then, for every agent, $ X $, there is a forgetting factor $ \gamma_{X} $. Due to the fact that every agent has two different tasks of blocking and teaching (the domain action and the communication action, respectively), we consider $ \gamma_{X} = \gamma_{1X}  + \gamma_{2X} $ according to Eq.~\ref{gamasamam}, where the first Gamma $ \gamma_{1X} $ stands for a forgetting factor in a blocking task belonging to the agent $ X $ and the second Gamma $ \gamma_{2X} $ refers to its forgetting factor in a teaching (helping) task. Furthermore, it is assumed that every agent can select to be absolutely selfish ($ \gamma_{1X} =0 , \gamma_{2X} = \gamma_{X}  $) as a defender, sacrifice its blocking task ($ \gamma_{1X} = \gamma_{X} , \gamma_{2X} = 0  $) to be more helpful in teaching, or have every other selection among these two border options, namely, $ \gamma_{X} = ( \gamma_{1X} , \gamma_{2X} ) $, $ \gamma_{1X} + \gamma_{2X} = \gamma_{X} $. Therefore, it can be noticed that a selfish strategy is a zero-cost communication for a given agent.  

Then, one can have
\begin{multline}
r_{max}^X (\alpha_{X} , \gamma_{1X}, \gamma_{2Y} ) =  \alpha_{X} \, p ( \gamma_{1X} ) + (1- \alpha_{X}) \left[ p ( \gamma_{1X} ) . p ( \gamma_{2Y} )  + q ( \gamma_{1X} ) . q ( \gamma_{2Y} ) \right] ,
\label{eqrmaxx}
\end{multline}\\[0.1cm]
according to Eq.~\ref{eqrmaxxx}.
For example, if 
\begin{equation}
\begin{cases}
\alpha_{X} = \alpha_{Y} = \alpha , \\
\forall k \quad \gamma_{kX} =  \gamma_{kY} = \gamma_{k} , 
\end{cases}
\Longrightarrow \quad r_{max}^X (\alpha , \gamma_{k}) = r_{max}^Y (\alpha , \gamma_{k}),
\end{equation}\\[0.1cm]
which refers to the symmetry of the problem between two agents. 
Otherwise, the efficiency of one agent would differ from that of its partner as a function of its own forgetting factor, its partner's forgetting factor and also, the parameter of observability of the environment for one agent. With the same variables of $ \gamma_{1X}, \gamma_{2Y} $, in Eq.~\ref{eqrmaxx}, one would have
\begin{multline}
r_{max}^Y (\alpha_{Y} , \gamma_{1X}, \gamma_{2Y} ) =  \alpha_{Y} \, p ( \gamma_{Y} - \gamma_{2Y} ) + (1- \alpha_{Y}) [ p ( \gamma_{Y} - \gamma_{2Y} ) . p ( \gamma_{X} - \gamma_{1X} ) \\ + q ( \gamma_{Y} - \gamma_{2Y} ) . q ( \gamma_{X} - \gamma_{1X} ) ]. 
\end{multline} 

\subsubsection{Classes of coalitions}
\label{superadditiven}
Considering the collective efficiency of two gaents as $ r_{col} = r_{max}^X + r_{max}^Y $, one may ask about the classes of coalitions (see \cite{book:MultiagentShoham} p. 386 for the definitions) in this game by comparing two cases: $ r_{col-FO} (\gamma_{X}, \gamma_{Y} ) $ in a fully observale environment versus $ r_{col-PO} (\alpha_{X},\alpha_{Y} , \gamma_{1X}, \gamma_{2Y}) $ in a partially observable environment. The former is straightforward as 
\begin{equation}
r_{col-FO} (\gamma_{X}, \gamma_{Y} )  = p ( \gamma_X ) + p ( \gamma_Y)
\end{equation} 
on the one hand. However, considering the latter, we deal with a four variable function. At first, we can see that every \textit{selfish-selfish} coalition $ \gamma_X = (0 , \gamma_X ) \,\, \& \,\, \gamma_Y = ( 0 , \gamma_Y ) $ will be superadditive, 
\begin{eqnarray}
r_{max}^X = \alpha_{X} + ( 1 - \alpha_X ) \, p ( \gamma_Y ) = p ( \gamma_Y ) +  \alpha_{X} \, q ( \gamma_Y ) , \notag \\
r_{max}^Y = \alpha_{Y} + ( 1 - \alpha_Y ) \, p ( \gamma_X ) = p ( \gamma_X ) +  \alpha_{Y} \, q ( \gamma_X ) . \notag 
\end{eqnarray}
Then,
\begin{equation}
r_{col-PO} (\alpha_X , \alpha_Y , \gamma_{1X} = 0 , \gamma_{2Y} = \gamma_{Y} ) \geq r_{col-FO} (\gamma_{X}, \gamma_{Y} ) .
\label{supsupadlll}
\end{equation}\\[0.1cm]
Equation~\ref{supsupadlll} will be satisfied for every $ 0 \leq \gamma_{X}, \gamma_{Y} \leq 1 $. Specifically, the underlying reason for being a superaddition in the \textit{selfish-selfish} coalition of $ \gamma_{X}= \gamma_{Y}= 1 $ is that it will increase the collective efficiency only due to the use of random information that two players share with each other; namely, the communication in this particular case is just sending an empty message with the zero-cost.     
 
Secondly, in the case of a \textit{sacrifice-sacrifice} coalition, $ \gamma_X = (\gamma_X , 0 ) \,\, \& \,\, \gamma_Y = (\gamma_Y , 0 ) $, we can obtain $ r_{max}^X  = p ( \gamma_X ) $, $ r_{max}^Y  = p ( \gamma_Y ) $ according to Eq.~\ref{eqrmaxx}, that is independent of $ \alpha_{X} \,\, \& \,\, \alpha_{Y} $. Therefore,
 \begin{equation}
 r_{col-PO} ( \alpha_{X}, \alpha_{Y}, \gamma_{1X} = \gamma_X , \gamma_{2Y}= 0 ) = r_{col-FO} (\gamma_{X}, \gamma_{Y} ).
\label{adlkntkacx}
 \end{equation}
Hence, every \textit{sacrifice-sacrifice} coalition will be an additive game.  
\begin{figure*}[h]
\centering
\subfigure[]{
\includegraphics[width=5.7cm, height=4.5cm]{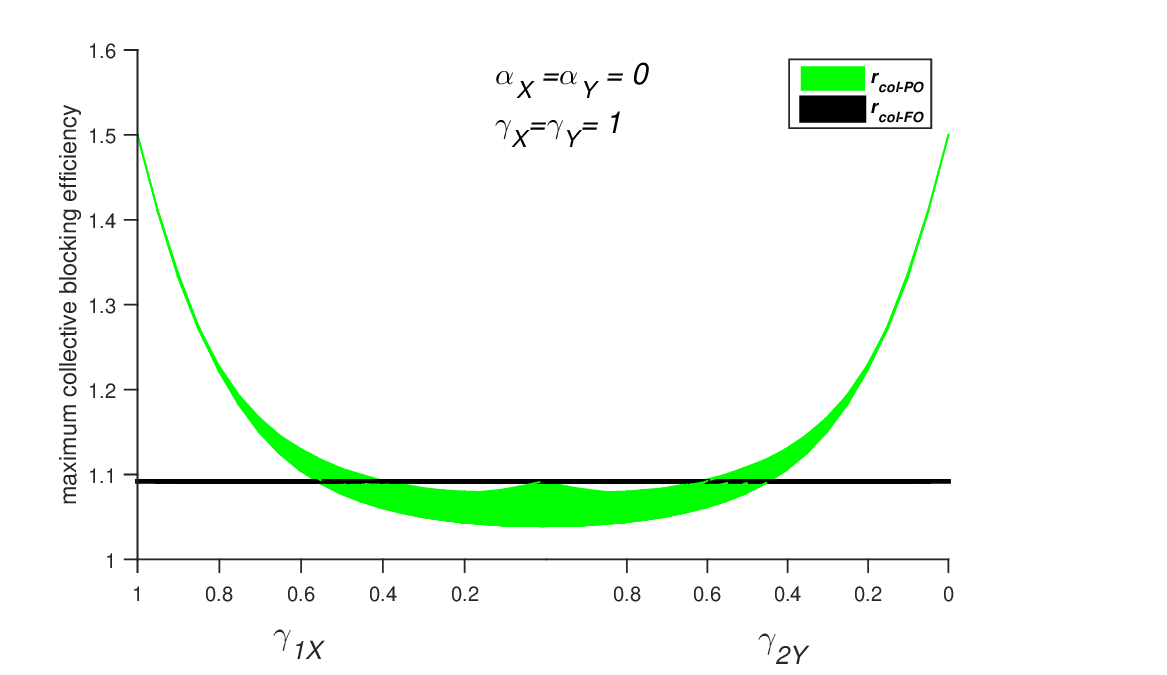}
\label{colclckeh0}
}
\subfigure[]{
\includegraphics[width=5.6cm, height=4.5cm]{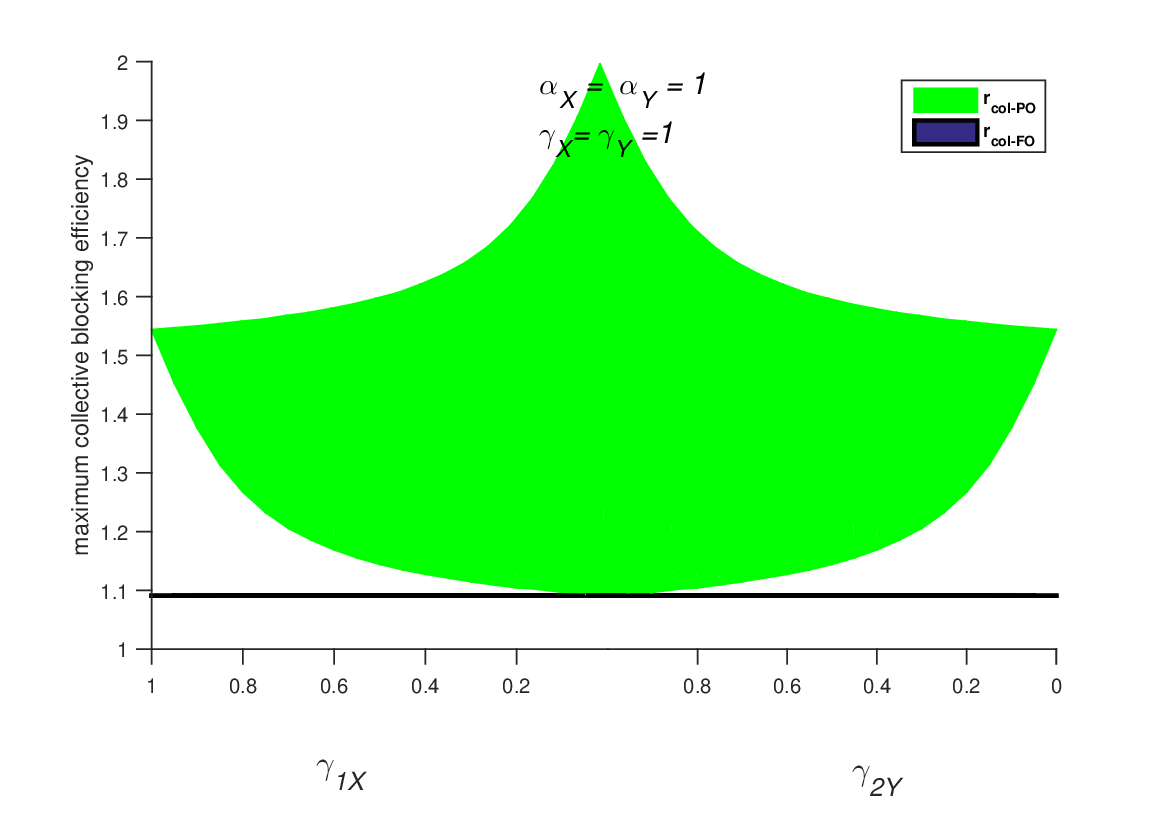}
\label{colcoclckeh1}
}
\caption{Maximum collective blocking efficiency in a 2-agent invasion game $ r_{col-PO} $ of Fig.~\ref{fig:1}, as compared with the maximum collective blocking efficiency in a fully observable environment $ r_{col-FO}  = p ( \gamma_X ) + p ( \gamma_Y) $ (the black line). The results have been illustrated for $ \gamma_X = \gamma_Y =1 $ and two cases of $ \alpha_{X} = \alpha_{Y} = \alpha $. Figure~\ref{colclckeh0} is for $ \alpha =0 $, where its maximums have accrued in the \textit{sacrifice-selfish} and \textit{selfish-sacrifice} selections, and Fig.~\ref{colcoclckeh1} is for $ \alpha =1 $, where the maximum happens in the \textit{selfish-selfish} selection}
\label{colcoclc}
\end{figure*}
On the contrary, while there are numerous selections for which the coalition is superadditive, especially in the big $ \alpha $ factors (see Fig.~\ref{colcoclckeh1} for $\alpha_{X} = \alpha_{Y} =1$), there are many other selections for which $ r_{col-PO} < r_{col-FO} $, according to the example of Fig.~\ref{colclckeh0}, especially for small observability parameters.  
\subsubsection{Equilibrium}
After all, we can consider the maximum blocking efficiencies of $ X $ and $ Y $ as a simple symmetric game. Furthermore, with $\gamma_{X} = \gamma_{Y} =1$, we can analyze the game using the results in Sec.~\ref{secsecdfga} and considering a small adjustment in the variables as 
\begin{equation}
\gamma_D \rightarrow \gamma_{1X}, \qquad \gamma_I \rightarrow \gamma_{2Y}, \qquad \alpha \rightarrow \alpha_X .
\label{moakasdi}
\end{equation}
regarding the agent $ X $.

Beside this, if we assume the two agents as two players, then every selection of every agent for its forgetting factor can be considered as a different pure strategy. For the condition of $ \gamma_{1X} + \gamma_{2X} = \gamma_{1Y} + \gamma_{2Y} = 1 $, a given pure strategy $ A $ would be a selection as $ A \equiv \gamma = [\gamma_{A}, 1-\gamma_{A}] $. Therefore, we can have two players for $ 2 $ different pure strategies $ A, B $ and build our game such that playing $ A [\gamma_{A}, 1-\gamma_{A}] $ for two agents means two agents have similar selections for their blocking and teaching forgetting factors. In addition, a payoff obtained by every agent could be considered as its maximum efficiency $ r_{max} (\alpha ) $ as a function of observability for every pure strategy. Having in mind that $ \alpha_{X} \neq \alpha_{Y}$, then in a general situation, there would be a variety of different payoffs for every agent. 

Nevertheless, regarding the modification of Eq.\ref{moakasdi}, considering some other restrictions might be helpful.
One can see that if every agent can choose between $ A ( \alpha_A ) $ and $ B ( \alpha_B ) $, then we can have a game as the following      
\begin{eqnarray}
 player2\,(Y) \quad \: \quad \nonumber \\ [0.2 cm]
player1 (X) \; \; \;
\begin{tabular}{|c|c|c|}
\hline  & $A ( \alpha_A )$ & $B ( \alpha_B )$ \\
\hline  $A ( \alpha_A ) $ & $a,a$ & $c,d$ \\
\hline  $ B ( \alpha_B )$ & $d,c$ & $b,b$ \\ 
\hline
\end{tabular}
\nonumber
\end{eqnarray}
\\[0.3cm]
, where the lowercase Latin letters refer to the playoffs of each player for a given pure strategy.
\begin{equation}
\begin{cases}
a = r_{max} ( \alpha_{A}, \gamma_{A}, 1 - \gamma_{A} ) \\
b = r_{max} ( \alpha_{B}, \gamma_{B}, 1- \gamma_{B} ) 
\end{cases} \\
,\qquad
\begin{cases}
c = r_{max} ( \alpha_{A}, \gamma_{A}, 1 - \gamma_{B} ) \\
d = r_{max} ( \alpha_{B}, \gamma_{B}, 1- \gamma_{A} ) .
\end{cases}
\end{equation}
Suppose that we always have $ \gamma_{A} <  \gamma_{B} $. Then, considering the modification of Eq.\ref{moakasdi}, Fig.~\ref{alpanim} as an example of $ \alpha = constant $ and tracking other plots in Fig.~\ref{alpamore} for other examples of $ \alpha = constant $ show that in a certain amount of $ \alpha $, we have $ a +c >  b + d $. Therefore, the strictly dominant strategy for every player is being absolutely selfish (selecting the lowest amount of the blocking forgetting factor), and since we have symmetry between the two agents, then (absolute) selfishness is also the only Nash equilibrium. It is a pure strategy strictly dominant for each player as well as Pareto optimal in the case of $ \alpha_A = \alpha_B $; it can also be considered as the fair share (the simplest Shapley value) between the agents in a superadditive coalition game.

In contrast, Fig.~\ref{alpamore} shows that regarding different amounts of the observability parameter, $ \alpha_{A} \neq \alpha_{B} $, one could have more complicated games. For example, the (absolute) selfishness as the Nash equilibrium can be dominated by another Pareto optimal solution which brings about a higher payoff to the two agents. For instance, tracking the scattering between the maximum and the minimum curves of $ r_{max} $ for $ \gamma_{1X} = 0 $ ($ \alpha_X = 0.99$ and $\alpha_X = 0 $, respectively) in Fig.~\ref{alpamore} shows that there would be a variety of other selections of $ \gamma_{1X} \neq 0 $ for which the related $ r_{max} $ has some intersections with that of (absolute) selfishness of $ \gamma_{1X} = 0 $ with $ \alpha_X =0 $. In other words, regarding different amounts of observability ($ \alpha_{A} \neq \alpha_{B} $), one can build other games such as a dilemma to be played as $ A( \alpha_A ) $ or $ B( \alpha_B ) $. Even more particularly, one can count for the payoffs if $ \gamma_A = 0 $, $ \gamma_B = 0.9 $ and $ \alpha_A = 0 $ and $ \alpha_B = 1$, leading to

\begin{eqnarray}
\begin{tabular}{|c|c|}
\hline  $p(1), p(1)$ & $p(0.1), p(0.9)$ \\
\hline  $p(0.9), p(0.1)$ & $p(0.9), p(0.9)$ \\ 
\hline
\end{tabular}
\nonumber
\end{eqnarray}
\\[0.3mm]
We can see that while $ p(1)$ is the Nash equilibrium, $ p(0.9) > p(1) $ is the Pareto optimal strategy; however, $ p(0.9) $ is not a Nash equilibrium itself.
\subsection{Multi-agent invasion game}
We bring a couple of 3-agent situated examples. The rest of the scenarios can be built upon a combination or modification of this two, as mentioned in Sec.~\ref{kaksobge2}. 

\begin{figure*}[h]
  \includegraphics[width=0.75\textwidth]{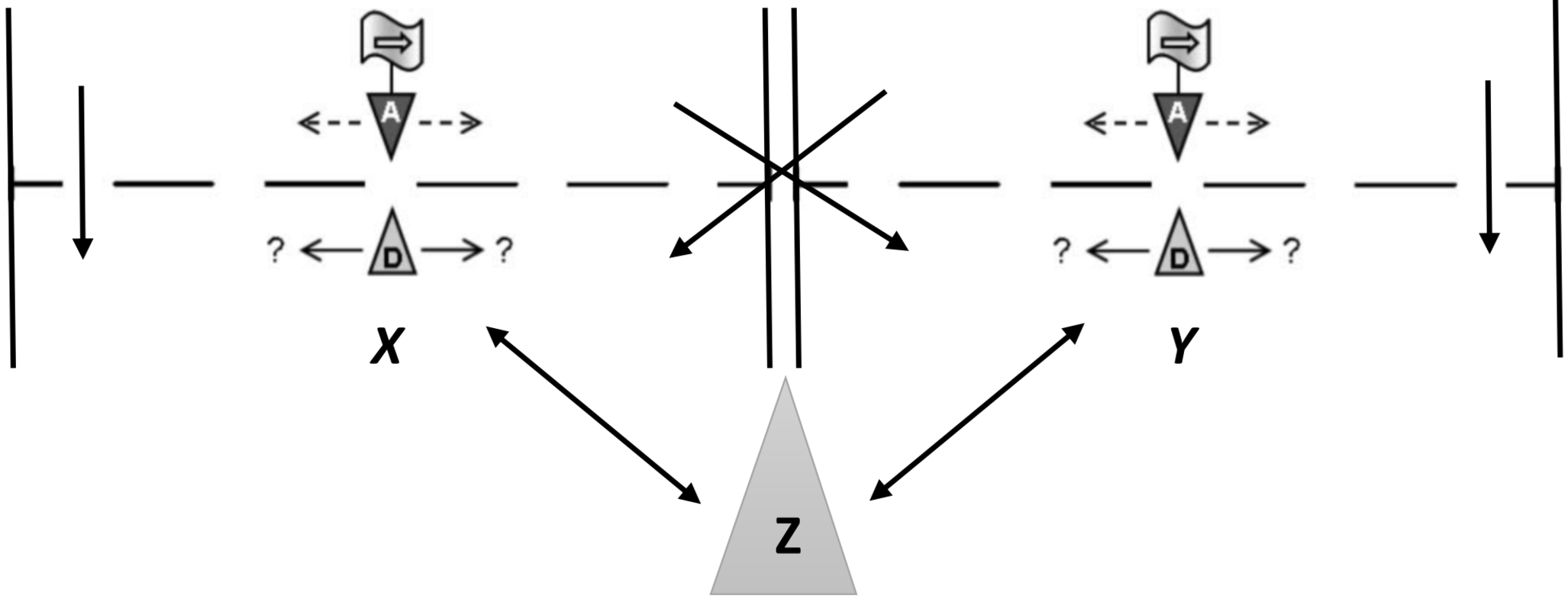}
\caption{A 3-agent invasion game where the agents $ X $ and $ Y $ are both the simultaneous protagonist-interpreter and $ Z$ is the second layer of interpretation}
\label{fig:fogs1}       
\end{figure*}

\begin{figure*}[h]
  \includegraphics[width=0.75\textwidth]{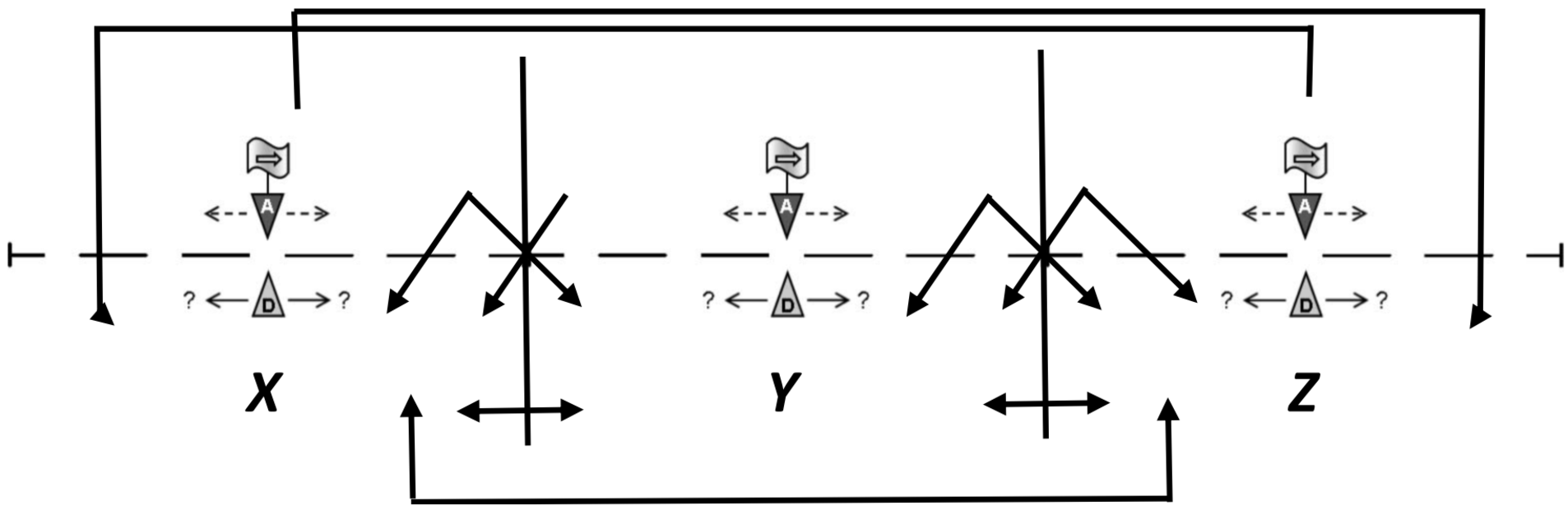}
\caption{A 3-agent invasion game where there are three protagonist-interpreter agents as $ X $, $ Y $ and $ Z $}
\label{fig:foggffogs1}       
\end{figure*}

On the one hand, Fig.~\ref{fig:fogs1} is a specific example from the invasion game based on Eq.~\ref{jankanjfl} and its clips network representation of Fig.~\ref{3agsmkdenall3}. 
On the other hand, Fig.~\ref{fig:foggffogs1} is a particular example of the invasion game based on Eq.~\ref{janflahspkan} and the related network of clips representation of Fig.~\ref{serid3agecps}.
     
\section{Conclusion}
In this paper, we added two new concepts of the \textit{belief projection operator} and the \textit{observability parameter} to the original scheme of projective simulation (PS) for the performance of an agent in a partially observable environment. In short, a given projector makes some belief states from the world states to the extent of the observability parameter of a given environment. Therefore, an individual policy for an agent can be defined as a map from the agent's belief state $ | (S+B)_{j} \rangle $ to an action $ | a_j  \rangle $. We used our method in a multi-agent setting where the projector was a second agent called an interpreter. Then we generalized our arrangement to include a system involving more than two agents. Although we expressed our results classically, our formulation with the modification of $ \langle a | s  \rangle \rightarrow {|\langle a | s  \rangle |}^{2} $ and their corresponding network representations could be used in the quantum regime too. 

For a defender, in our invasion example, the fully observable percepts were coming directly from the attacker while the partially observable ones were conveying indirectly through the interpreter. We continued our case example to include a 2-agent toy problem, where, on one hand, the selfishness selection regarding a two-part forgetting factor could be considered as a zero-cost communication option giving rise to a superadditive coalition. On the other hand, the selfishness could be considered as a Nash equilibrium, though it could be dominated by another Pareto optimal strategy in case $ \alpha_{i} \neq \alpha_{j} $. One can distinguish between the belief states and the world states by adding another property, such as different colors, to them for simplicity. That is, for example, $ | s\rangle \in \{ | \Leftarrow \rangle , | \Rightarrow \rangle \} $ and $ | b\rangle \in \{ | { \color{mygreen} \Leftarrow} \rangle , | {\color{mygreen} \Rightarrow} \rangle \} $ for $ N=2 $ in a partially observable invasion toy problem. However, it is important to note that the additional properties such as different colors do not play a significant role in the theory of partial observability of Eq.~\ref{jankan}. In fact, the underlying difference between the original PS (a compositional memory added to an episodic memory) and the current study is the difference between the forgetting factor of an episodic memory and the forgetting factor of a compositional memory, here ($ \gamma_{pa} \neq \gamma_{I} $). At the same time, as shown in Appendix.~\ref{erageffredg}, there will be some limitations as using "edge glow" to expand a single-agent PS learning method toward a multi-agent approach.

Although the system state remained jointly fully observable, there was not a joint reward function in our formulation to build a decentralized approach. Instead, there has been selected a self-interested perspective, similar to the existing interactive literature. However, the method needs further works to create a generalized belief state regarding some beliefs including agents' parameters as well as the environmental parameters to build either decentralized or interactive schemes depending on the situation. It might be done utilizing a meta-learning \cite{Briegel3} or other optimal solutions with respect to the amounts of forgetting factors, etc. in the specific problems. As might be expected, other works can be done considering centralized and decentralized approaches including sharing clips, sharing rewards or sharing policies, etc. among some cooperating agents such as multi-agent grid-world tasks mentioned in \cite{tantann} or tiger toy problem \cite{Nair,Nair2}. 

Moreover, there could be an interesting quantum consideration too. This is because some clips in the networks belong to different agents that can be spatially separate from the other clips, which require the notion of remote entanglement in the quantum context. Similarly, there are some approximate branches in literature such as distributed quantum computation and quantum game theory \cite{Buhrmanh,Perseguersk,Meyer,Miakkisz,Psiotrowlski,fFlitney,saiff}.   

Nonetheless, in my opinion, the PS model and its partially observable method can be considered widely in psychology or behavioural  economics too. Owing to an assumptive difference between the fictitious memory clips of $\textcircled{\textit{s}}$, $\textcircled{\textit{a}}$ and their actual counterparts ($s, \, a$), the approach might be a compelling context for understanding the various decision-making processes among artificial people in a certain condition via different perceptions of a specific situation. For example, merging some memory clips in some compositional memory may be applied to build an abstract clip network which can be related to a creative thinking or an illusion related to one type of personality as "openness to new experiences" \cite{AAntinori} in the five-factor model, which we might consider as another study. 

Further, the partially observable PS might be utilized in psychology by itself. Because a projector may not be just an exterior interpreter; instead, it can be considered as an interior brain structure of a given agent. For instance, a brain can have some projections from the childhood; as a result, it touches on the perception of a given situation and therefore, affects the performance or the decision-making processes. Consequently, an internal projection as a part might bring about some subconscious notion among individuals or even different cultures among societies via some more general interior-exterior projectors.

Finally, as we saw, the transparency was dependent on three parameters, the environment parameter $ \alpha $, the agent parameter $ \gamma_{pa} $ and the belief states affected by another agent, $ \gamma_{I} $. In a general perspective, after all, the transparency or a given perception can be dependent on the environment, the agent itself, and its society.           
  
\begin{acknowledgements}
I thank Alexey Melnikov for our long discussions. I am also grateful for the rejection comments offered by an editor from 
Autonomous Agents and Multi-Agent Systems journal regarding the previous version. I think the final version could not have been completed without some illuminating feedbacks. The author also acknowledges H. Bassereh, V. Salari, and M. Ghadimi for their help.
\end{acknowledgements}

\appendix

\section*{Appendix 1: Actions versus the probability of actions}\label{averagefequa}
While in the original papers of PS, the averaged performing rewarded actions are depicted for the efficiency, we used the probability of doing rewarded actions $ r^{(t)} $ for the same purpose.
\begin{figure*}[h]
\centering
\subfigure[]{
\includegraphics[width=5.6cm, height=4.7cm]{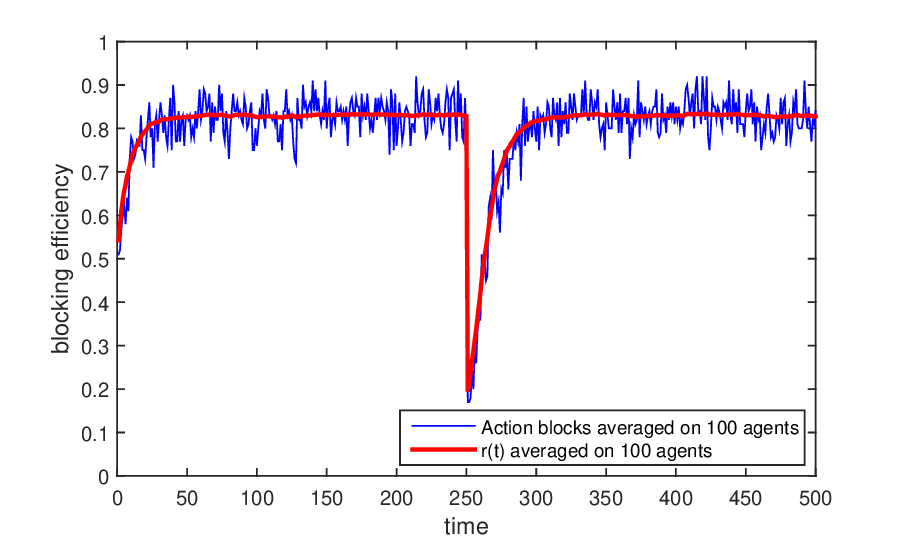}
\label{figappendix1}
}
\subfigure[]{
\includegraphics[width=5.6cm, height=4.7cm]{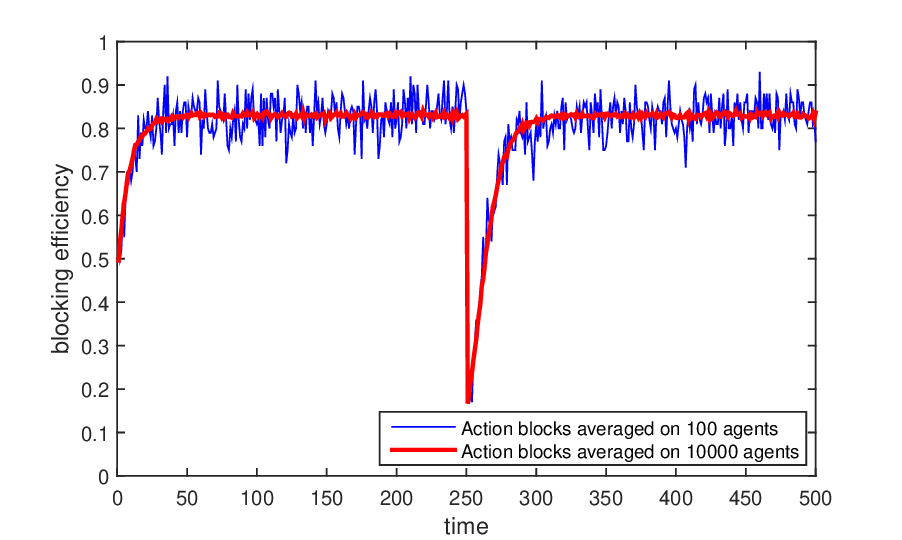}
\label{figappendix2}
}
\caption{Blocking actions (blue) versus the probability of blocking, $ r^{(t)} $ of Eq.~\ref{eqoriggin}, in a fully observable and changing environment for the one agent invasion game ($ N=2 $). Figures are depicted for $ \gamma=0.1 $}
\label{figappendix11}
\end{figure*}
It is due to the fact that the probability of doing an action in a large time step $ t $, as in Fig.~\ref{figappendix11}, is a better approximation for $ r^{( t \to \infty )} $ than the real actions in $ t $.  
\section*{Appendix 2: Reward function for the 2-agent invasion and the edge glow}\label{erageffredg}
The reward function $ \lambda $ for a pair of percept-action in a fully observable one agent scenario is defined as
\begin{equation}
\begin{cases}
\lambda ( s_i , a_k ) > 0 , \quad i= k \\
\lambda ( s_i , a_k ) = 0 , \quad i \neq k . 
\end{cases}
\end{equation}
For a multi-agent approach, on the other hand, we might have different options. However, we have to know that our reward definition shouldn't destroy the partial observability or other agents' contribution that is in opposition to the purpose.     

\begin{figure*}[h]
\centering
\subfigure[]{
\includegraphics[width=4.5cm, height=5cm]{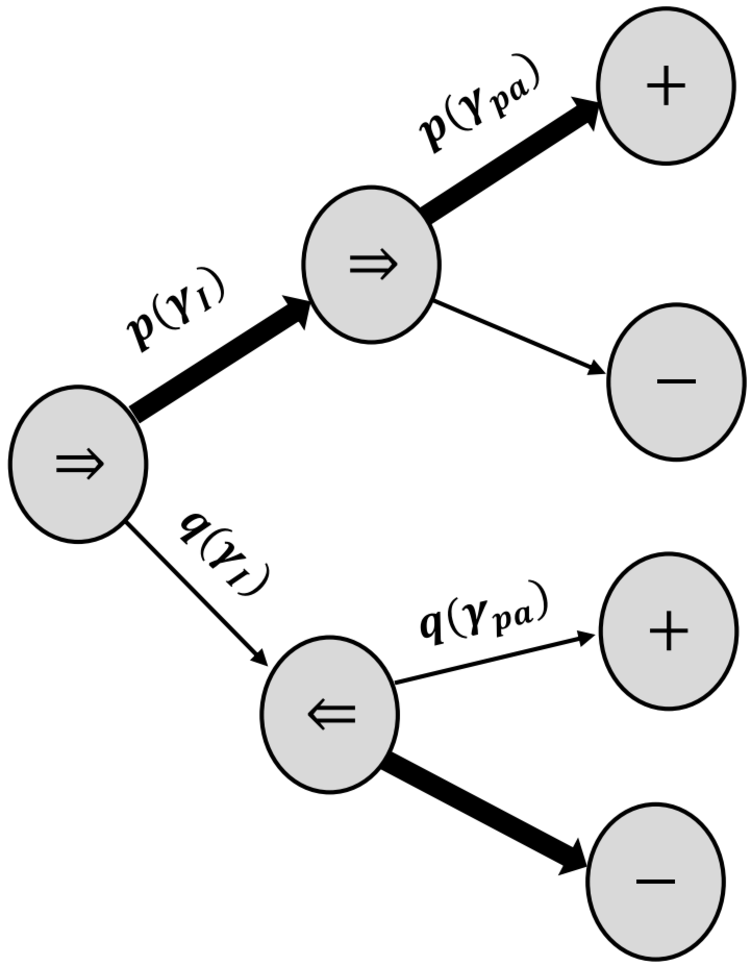}
\label{figag11ndix1}
}
\hspace*{1.5cm}
\subfigure[]{
\includegraphics[width=4.5cm, height=5cm]{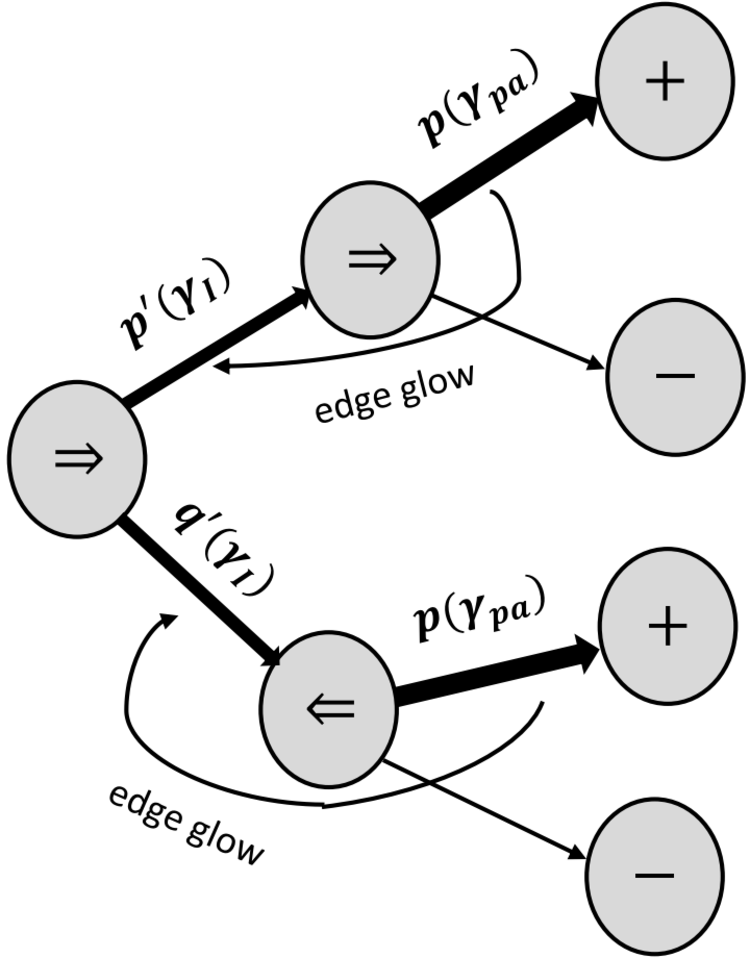}
\label{figapg2dix2}
}
\caption{Two given reward schemes for one of the world percepts ($ \Rightarrow $) of the interactive invasion game. Fig.~\ref{figag11ndix1} is depicted for the reward functions as defined in Eq.\ref{eqkllberahme}, and Fig.~\ref{figapg2dix2} is (a wrong option) related to utilizing an edge glow in case the rewards are given to the blocking actions}
\label{figappsdlix1}
\end{figure*}

For our solved interactive invasion problem of the paper, I used the rewards as
\begin{equation}
\begin{cases}
\lambda ( s_i , \, a_k ) = 1 , \quad i= k \\
\lambda ( s_i , a_k ) = 0 , \quad i \neq k 
\end{cases},
\begin{cases}
\lambda ( s_i , \, b_j ) = 1 , \quad i= j \\
\lambda ( s_i , b_j ) = 0 , \quad i \neq j 
\end{cases},
\begin{cases}
\lambda ( b_j , \, a_k ) = 1 , \quad j= k \\
\lambda ( b_j , a_k ) = 0 , \quad j \neq k .
\end{cases}
\label{eqkllberahme}
\end{equation}
The relevant network representation for one of the world percepts ($ \Rightarrow $) is shown in Fig.~\ref{figag11ndix1}.

One may suppose that employing a reward for the blocking actions and using the ''edge glow'' \cite{Briegel2} could speed up learning as depicted in Fig.~\ref{figapg2dix2}. However, such a bypass reward option is omitting the interpreter as another learning agent. Because 
\begin{eqnarray}
p ( \gamma_{pa} ) . p ( \gamma_{I} )  + q ( \gamma_{pa} ) . q ( \gamma_{I} ) \, \leq \, p ( \gamma_{pa} ) 
\label{yajsatl}
\end{eqnarray}
in Fig.~\ref{figag11ndix1}, which refers to the partial observability of the environment for the protagonist agent as indicated in Eq.~\ref{eq2020} (the equality sign in Eq.~\ref{yajsatl} is preserved for $ \gamma_{I} = 0 $). But, 
\begin{equation}
p ( \gamma_{pa} ) . p' ( \gamma_{I} )  + p ( \gamma_{pa} ) . q' ( \gamma_{I} ) = p ( \gamma_{pa} ) [ p' ( \gamma_{I} )  + q' ( \gamma_{I} ) ] = p ( \gamma_{pa} ) 
\end{equation}
in Fig.~\ref{figapg2dix2}, by which the partial observability and the role of the interpreter will be destroyed completely. As a result, we have chosen Fig.~\ref{figag11ndix1} and Eq.~\ref{eqkllberahme} for the interactive reward function.

\end{document}